\documentclass[11pt]{article}

\usepackage[preprint]{acl}

\usepackage{times}
\usepackage{latexsym}

\usepackage[T1]{fontenc}
\usepackage[utf8]{inputenc}

\usepackage{microtype}

\usepackage{inconsolata}

\usepackage{graphicx}

\usepackage{hyperref}       % hyperlinks
\usepackage{url}            % simple URL typesetting
\usepackage{booktabs}       % professional-quality tables
\usepackage{amsfonts}       % blackboard math symbols
\usepackage{nicefrac}       % compact symbols for 1/2, etc.
\usepackage{microtype}      % microtypography
\usepackage{xcolor}         % colors
\usepackage{graphicx}
\usepackage{algorithm}      
\usepackage{algpseudocode}  
\usepackage{caption}
\usepackage{amsmath}
\usepackage{threeparttable}
\usepackage{colortbl}
\definecolor{oursbg}{RGB}{230, 245, 230}
\usepackage[dvipsnames]{xcolor}
\usepackage{adjustbox}
\usepackage{enumitem}
\usepackage{tabularx}    % tabularX 环境
\usepackage{array}       % >{\centering\arraybackslash}
\usepackage[most]{tcolorbox}
\usepackage{float}
\usepackage{amssymb}
\usepackage{subcaption}
\usepackage{fontawesome5}

\title{When to Vote, When to Rewrite: Answer Stability Guides Test-Time Scaling}

\author{
  Zhimin Lin$^{1}$,
  Yixin Ji$^{1}$,
  Jinpeng Li$^{2}$,
  Yu Luo$^{2}$,
  Dong Li$^{2}$,
  \textbf{Min Zhang}$^{3}$\thanks{\; Corresponding author.} \\
  $^{1}$School of Computer Science and Technology, Soochow University\\
  $^{2}$Department of Foundation Model, 2012 Labs, Huawei \\
  $^{3}$Harbin Institute of Technology, Shenzhen (HITSZ)\\
  \texttt{\{linzhimin327, jiyixin169\}@gmail.com} \\
  \faGithub\ \url{https://github.com/NenRinCake/PIVOT_Stability-Guided-Routing}
}

\begin{document}
\maketitle
\begin{abstract}
Test-time scaling improves the reasoning performance of Large Reasoning Models (LRMs), but the appropriate use of additional computation varies across instances. We identify a stability-dependent complementarity: majority voting is more effective on answer-stable instances, whereas problem rewriting is more effective on answer-unstable instances. We propose \textsc{PIVOT}, a training-free sequential routing framework that uses persistent disagreement to guide the switch from candidate aggregation to problem reformulation. It stops sampling upon agreement and invokes rewriting followed by a single solve when disagreement persists through the check budget. Generated answers are reused for strategy selection and final answering. Experiments on seven mathematical benchmarks across three models show approximately 3--7-point gains in average accuracy over the strongest baselines. Efficiency evaluations show approximately 50\% lower token usage than six-sample majority voting. \textsc{PIVOT} even surpasses the empirical oracle-selection ceiling of six original-formulation samples, with gains on code generation.
\end{abstract}

\section{Introduction}

Test-time scaling has made inference-time computation an important resource for improving the reasoning capabilities of Large Reasoning Models (LRMs) \citep{wang2022self,yao2023tree,lin2024justasktimeselfagreement}. Yet the value of additional computation depends not only on how much is spent, but also on the solution distribution induced by the current problem representation. Repeated sampling and majority voting aggregate evidence from this distribution \citep{wang2022self,li2024agentsneed,gui2024bonbon}, whereas problem rewriting changes the representation from which solutions are generated \citep{zhou2024paraphrase,zhang2025rm,lai2025multidimensional}. This raises a fundamental question: \textit{when should additional computation exploit the current solution distribution, and when should it instead reshape it?}

\begin{figure}[t]
  \includegraphics[width=\columnwidth]{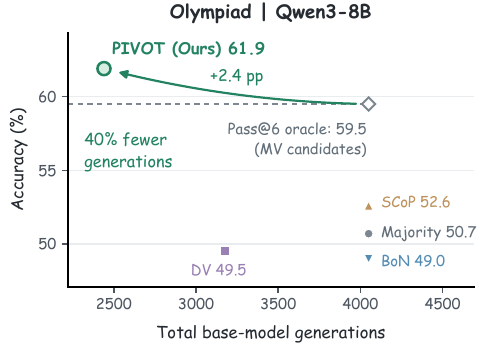}
  \caption{\textbf{More with Less.} Accuracy and generation cost on OlympiadBench with Qwen3-8B.}
  \label{fig:more_with_less}
\end{figure}

\begin{figure*}[t]
  \includegraphics[width=\textwidth]{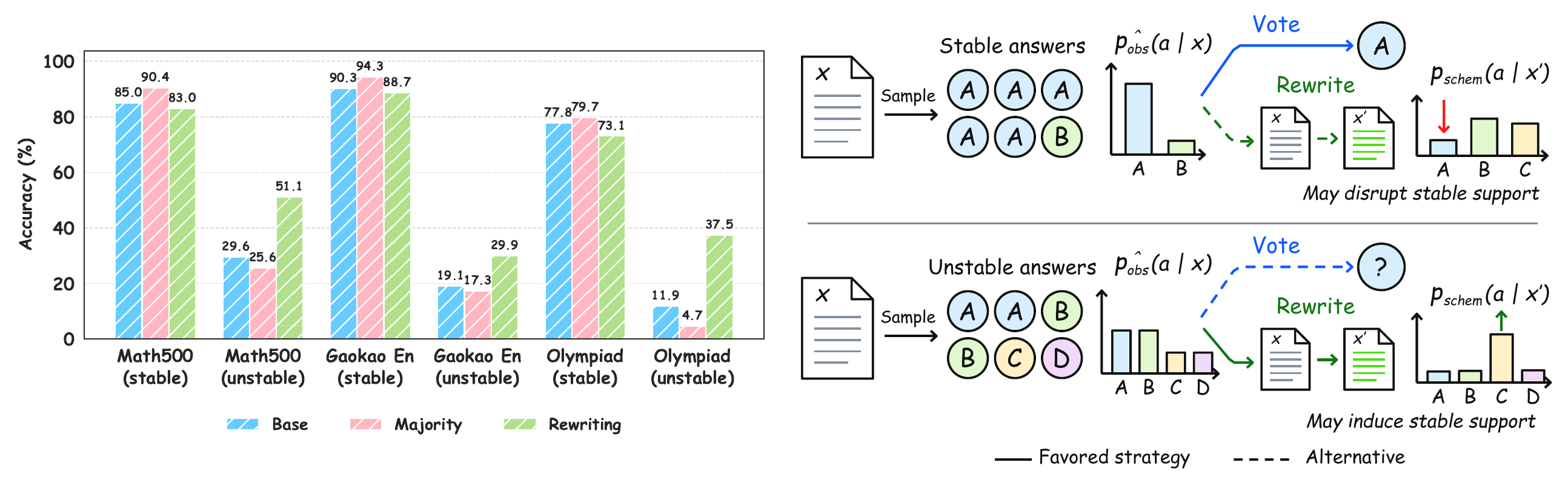}
  \caption{Complementarity between majority voting and rewriting. \textbf{Left:} Strategy accuracy on stable and unstable subsets, where stability requires at least five of six sampled answers to agree. \textbf{Right:} A conceptual illustration of how voting exploits existing answer support and rewriting changes it. $\hat p_{\mathrm{obs}}$ denotes empirical frequencies, and $p_{\mathrm{schem}}$ denotes a schematic post-rewrite distribution.}
  \label{fig:motivation}
\end{figure*}

We investigate whether answer stability can guide the choice between candidate aggregation and problem reformulation by comparing the relative performance of majority voting and rewriting across stability-defined subsets. We classify an instance as stable if at least five of six sampled answers agree, and unstable otherwise. As shown in Figure~\ref{fig:motivation} (left), \textbf{the relative advantage of the two strategies reverses across these subsets}: majority voting outperforms rewriting on stable subsets, whereas rewriting outperforms majority voting on unstable subsets across all three benchmarks. Rewriting also reduces accuracy relative to base sampling on stable subsets but improves it on unstable subsets. This complementarity suggests that answer stability may help determine whether additional computation should aggregate existing candidates or change the problem formulation.

A possible explanation lies in how the two strategies interact with answer support, as illustrated in Figure~\ref{fig:motivation} (right). Majority voting exploits repeated support among candidates generated under the same problem representation. Concentrated support provides a stronger basis for selecting a dominant answer, consistent with the higher voting accuracy observed on stable subsets; rewriting may instead disrupt this established support. In contrast, dispersed empirical support suggests that the current representation may not induce a solution distribution with stable answer support. Rewriting reformulates the problem in an attempt to induce such a distribution and may therefore be more beneficial than further aggregation.

However, reliably characterizing the answer-support distribution requires repeated sampling, potentially offsetting the computational benefits of strategy selection. Agreement among independently sampled answers is more likely when answer support is concentrated, making disagreement a lightweight proxy for stability. This signal is directly available from generated answers and requires neither additional training nor an external evaluator. Since initial disagreement may reflect sampling variability, an additional check provides an opportunity to observe emerging agreement. Persistent disagreement indicates that these checks have not established consistent answer support, motivating a switch from aggregation to reformulation.

Building on this observation, we introduce \textsc{PIVOT}, a training-free sequential routing framework that uses persistent disagreement to guide the switch from candidate aggregation to problem reformulation. It reuses sampled answers for strategy selection and final answering, terminating upon agreement and invoking rewriting followed by a single solve when disagreement persists through the check budget.

Our main contributions are:
\begin{itemize}
    \item We identify a stability-dependent reversal in the relative performance of majority voting and problem rewriting, motivating instance-level allocation between candidate aggregation and reformulation.
    \item We develop a training-free sequential routing framework that jointly determines when to stop sampling and when to switch from candidate aggregation to problem reformulation. By reusing answers to assess whether disagreement persists, it enables adaptive strategy selection without a separate stability-estimation stage or an external verifier.
    \item On mathematical reasoning benchmarks, our framework achieves 3--7-point gains in average accuracy with approximately half the token usage of six-sample majority voting, and exceeds the empirical original-formulation pass@6 oracle (Table~\ref{tab:pass6_oracle}). Additional gains on code generation demonstrate its applicability beyond mathematics.
\end{itemize}

\begin{figure*}[t]
  \centering
  \includegraphics[width=\textwidth]{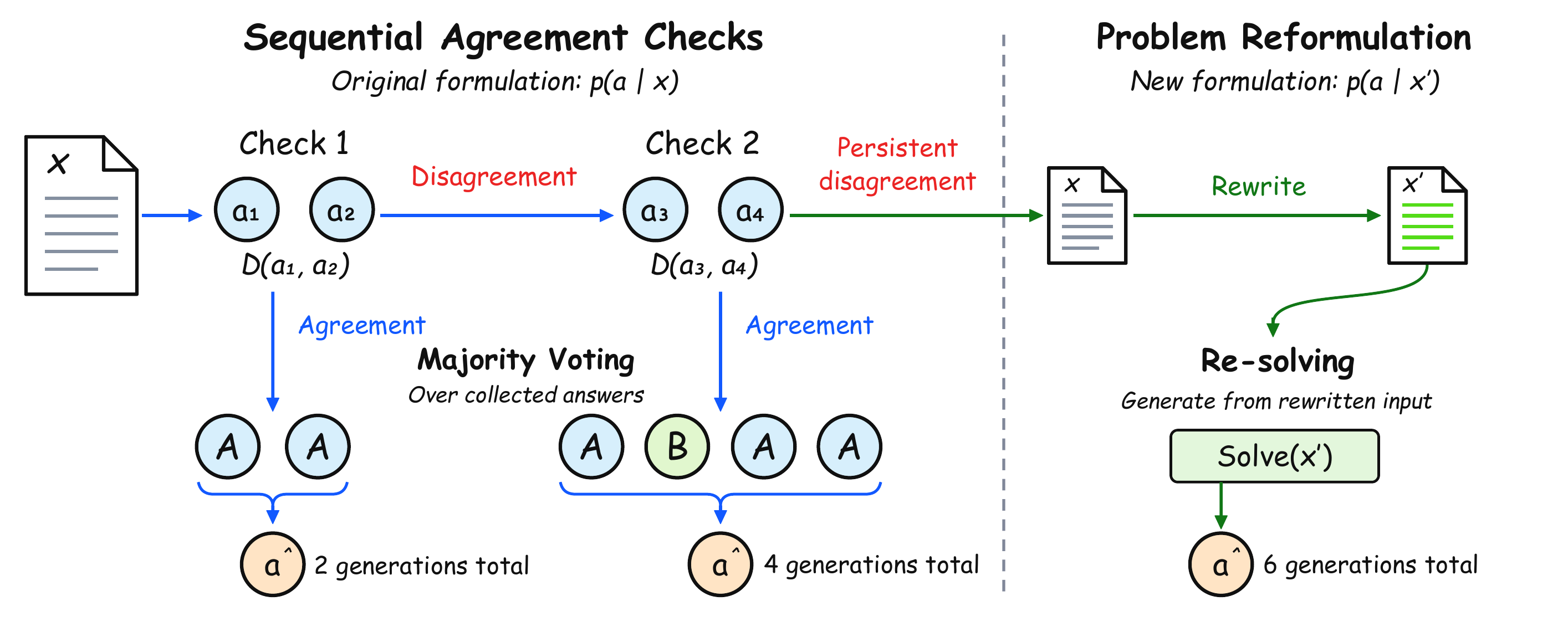}
  \caption{Overview of sequential strategy selection with the default two-check configuration. Agreement at either check stops sampling and selects the final answer by majority voting over collected answers. Persistent disagreement triggers problem reformulation followed by a single solve. $D$ denotes the pairwise disagreement indicator defined in Equation~\ref{eq:disagreement_check}. Generation counts are cumulative and include rewriting and re-solving.}
  \label{fig:pipeline}
\end{figure*}

\section{Related Work}

\subsection{Test-Time Scaling and Adaptive Computation}

Test-time scaling improves model performance through additional inference-time computation. Representative approaches include Self-Consistency \citep{wang2022self}, which aggregates sampled reasoning paths; Best-of-N sampling \citep{gui2024bonbon}, which selects high-scoring candidates; iterative self-correction \citep{shinn2023reflexionlanguageagentsverbal,chen2023teachinglargelanguagemodels,kumar2024training}; and search over reasoning paths \citep{yao2023tree,zhang2024accessing,bi2024forest,gao2024interpretable}. Efficiency-oriented methods further investigate when additional computation is necessary. Adaptive-Consistency \citep{aggarwal2023letssamplestepstep} and Dynamic Voting \citep{xue2023dynamic} use agreement among sampled answers to adjust sampling and determine when to stop. \citet{xiang2026thinkingenoughearlyexit} study early exit through sufficiency assessment for efficient reasoning. Speculative Rejection \citep{zhang2024accelerating} accelerates Best-of-N by terminating candidates that are unlikely to receive the highest final reward. Other studies examine the linguistic generalizability of test-time scaling \citep{son2025linguisticgeneralizabilitytesttimescaling} and its limitations on knowledge-intensive tasks \citep{zhao2026testtimescalingreasoningmodels}. Our work examines whether answer stability can guide the allocation of computation between candidate aggregation and problem reformulation.

\subsection{Problem Reformulation for Reasoning}

Problem reformulation changes the input from which a language model generates solutions. In broader text rewriting, RewriteLM \citep{shu2024rewritelm} uses instruction tuning and reinforcement learning to improve instruction-following rewriting. For mathematical reasoning, SCoP \citep{zhou2024paraphrase} investigates the positive and negative effects of surface-form changes and distributes reasoning samples across paraphrases before aggregating their answers through self-consistency. RM-PoT \citep{zhang2025rm} integrates mathematical problem reformulation, domain-aware example retrieval, and executable code generation. Multidimensional Consistency \citep{lai2025multidimensional} varies the order of in-context examples, problem phrasing, and language, and aggregates answer consistency across these dimensions to improve mathematical reasoning. Reformulation has also been studied for selective answering in factual question answering. The question-level probing framework of \citet{bin2026knowledge} investigates whether questions rejected under an unstable original prompt can be recovered through semantically equivalent reformulations. It checks reformulation equivalence and requires consistent low-uncertainty evidence across multiple views before accepting an answer, improving the risk--coverage trade-off.

\section{Methodology}
\label{sec:methodology}

\textsc{PIVOT} follows a simple design principle: use inexpensive observations of answer agreement to guide the allocation of computation between candidate aggregation and problem reformulation. We instantiate this principle with a two-check sequential policy that allows early termination and invokes rewriting only when disagreement persists across both checks.

\subsection{Answer Stability and Disagreement}
\label{sec:cate}

Given a problem $x$, let $p(a \mid x)$ denote the distribution over normalized final answers induced by the model under a fixed reasoning prompt and sampling configuration. Answer support describes how probability mass is distributed across these answers. Our motivation analysis shows that the relative performance of candidate aggregation and problem reformulation varies with answer stability. Since directly estimating the answer distribution requires repeated sampling, we use pairwise agreement as a lightweight observation of this stability.

For two independently sampled answers $a_1$ and $a_2$, the probability of agreement is
\begin{equation}
\Pr(a_1 = a_2 \mid x)
= \sum_a p(a \mid x)^2.
\label{eq:agreement_probability}
\end{equation}
This quantity measures the concentration of answer support: more concentrated distributions yield agreement more frequently. Agreement therefore provides evidence of answer stability, rather than a guarantee of correctness.

We define pairwise disagreement as
\begin{equation}
D(a_i,a_j) = \mathbb{I}[a_i \not\equiv a_j],
\label{eq:disagreement_check}
\end{equation}
where $\equiv$ denotes answer equivalence under the normalized matching procedure described in Appendix~\ref{sec:equivalence}, accounting for numerical, structural, and symbolic equivalence. The answer categories in Equation~\ref{eq:agreement_probability} follow the same equivalence procedure.

A single disagreement may arise from sampling variability, even when answer support is concentrated. We therefore use sequential checks on independently sampled pairs to assess whether disagreement persists with additional sampling. Each check uses two answers, the minimum required to observe disagreement. Table~\ref{tab:easy_recall} shows that requiring more answers to agree reduces the recall of correct instances, further motivating a lightweight pairwise check. Additional checks reduce the chance of triggering reformulation solely due to an isolated disagreement, but incur further sampling cost. The number of checks is configurable, balancing evidence for strategy selection against inference cost. We adopt two checks as the default configuration and evaluate one- and three-check variants in Table~\ref{tab:sampling_budget}. The six-sample stability criterion in Figure~\ref{fig:motivation} defines the diagnostic subsets only; inference uses these sequential pairwise checks.

\subsection{Sequential Strategy Selection}
\label{sec:filter}

At each round, the model generates two independent answers under the original problem formulation. Agreement stops sampling, and the final answer is selected by majority voting over all collected answers, with ties broken at random. Otherwise, another pair is sampled while the check budget permits; persistent disagreement across all permitted checks triggers problem reformulation.

Under the default two-check configuration, the first round produces $a_1$ and $a_2$. If $D(a_1,a_2)=0$, their shared answer is returned. Otherwise, the model samples a second independent pair, $a_3$ and $a_4$. If $D(a_3,a_4)=0$, the initial disagreement is transient, and the final answer is selected by voting over all four candidates. Since the first pair disagrees, the answer shared by the second pair necessarily determines the voting result. If $D(a_1,a_2)=D(a_3,a_4)=1$, disagreement persists across both checks, and the policy invokes problem reformulation followed by re-solving.

\subsection{Problem Reformulation and Re-solving}
\label{sec:rewriting}

For instances with persistent disagreement, the model rewrites the original problem $x$ into a new formulation $x'$ using the prompt in Appendix~\ref{appendix:prompts}. The prompt instructs the model to simplify the description while preserving its meaning, important quantities, and constraints, and to return only the rewritten problem without answers or calculations.

Rewriting changes the input from which solutions are generated, shifting the answer distribution from $p(a\mid x)$ to $p(a\mid x')$. Rather than aggregating further candidates under the original formulation, this intervention attempts to induce a solution distribution with stable answer support. The model then solves $x'$ once and returns the resulting answer as the final answer. A single model performs all original-problem solves, rewriting, and re-solving, with the same reasoning prompt used for both original and rewritten problems. No additional sampling or voting is performed after rewriting.

Under the default two-check configuration, agreement at the first or second check yields a final answer after two or four generations, respectively. The reformulation route requires six generations: four original-problem solves, one rewrite, and one solve of the rewritten problem.
\begin{table*}[t]
\centering

\small
\setlength{\tabcolsep}{2.5pt}
\begin{threeparttable}
\begin{adjustbox}{width=\linewidth,center}
\begin{tabular}{lcccccccc}
\toprule
\textbf{Model / Method} & \textbf{Math500} & \textbf{Gaokao En} & \textbf{Olympiad} & \textbf{GSM8K} & \textbf{AMC23} & \textbf{AIME25} & \textbf{AIME24} & \textbf{Avg} \\
\midrule
Qwen3-8B & 84.4 & 77.7 & 46.4 & 95.7 & 65.0 & 23.3 & 36.7 & 61.3{\tiny $\pm$0.71} \\
Qwen3-8B (Majority) & 87.6 & 79.7 & 50.7 & 95.8 & 85.0 & 33.3 & 50.0 & 68.9{\tiny $\pm$0.26} \\
Qwen3-8B (DV) & 87.6 & 79.7 & 49.5 & 95.7 & 85.0 & 30.0 & 43.3 & 67.3{\tiny $\pm$0.33} \\
Qwen3-8B (SCoP) & 88.2 & 79.5 & 52.6 & 95.4 & 82.5 & 33.3 & 40.0 & 67.4{\tiny $\pm$0.34} \\
Qwen3-8B (BoN) & 85.8 & 75.6 & 49.0 & 95.1 & 65.0 & 33.3 & 40.0 & 63.4{\tiny $\pm$0.57} \\
\rowcolor{oursbg} Qwen3-8B (Ours) & \textbf{92.8} & \textbf{82.9} & \textbf{61.9} & \textbf{95.8} & \textbf{87.5} & \textbf{43.3} & \textbf{66.7} & \textbf{75.8{\tiny $\pm$0.43}} \\
\midrule
Qwen3-4B & 85.4 & 75.6 & 49.0 & 94.2 & 65.0 & 23.3 & 30.0 & 60.4{\tiny $\pm$0.62} \\
Qwen3-4B (Majority) & 88.6 & 79.5 & 53.2 & \textbf{95.5} & 82.5 & 26.7 & 50.0 & 68.0{\tiny $\pm$0.27} \\
Qwen3-4B (DV) & 88.6 & 78.4 & 51.3 & 95.1 & 80.0 & 30.0 & 46.7 & 67.2{\tiny $\pm$0.31} \\
Qwen3-4B (SCoP) & 87.8 & 80.3 & 55.6 & 93.5 & 82.5 & 26.7 & 40.0 & 66.6{\tiny $\pm$0.36} \\
Qwen3-4B (BoN) & 87.4 & 76.4 & 50.1 & 94.1 & 72.5 & 26.7 & 50.0 & 65.3{\tiny $\pm$0.59} \\
\rowcolor{oursbg} Qwen3-4B (Ours) & \textbf{90.6} & \textbf{81.8} & \textbf{57.9} & 94.5 & \textbf{87.5} & \textbf{43.3} & \textbf{56.7} & \textbf{73.2{\tiny $\pm$0.55}} \\
\midrule
DS-Llama-8B & 80.2 & 74.0 & 46.4 & 85.0 & 67.5 & 30.0 & 40.0 & 60.4{\tiny $\pm$0.80} \\
DS-Llama-8B (Majority) & 84.6 & 79.8 & 51.6 & 87.5 & 80.0 & \textbf{33.3} & 46.7 & 66.2{\tiny $\pm$0.42} \\
DS-Llama-8B (DV) & 85.8 & 76.9 & 51.0 & 87.5 & 77.5 & \textbf{33.3 }& 46.7 & 65.5{\tiny $\pm$0.43} \\
DS-Llama-8B (SCoP) & 88.2 & 78.2 & 51.9 & 89.0 & 85.0 & \textbf{33.3} & 40.0 & 66.5{\tiny $\pm$0.53} \\
DS-Llama-8B (BoN) & 79.0 & 78.8 & \textbf{55.3} & 88.8 & 75.0 & 30.0 & 43.3 & 64.3{\tiny $\pm$0.76} \\
\rowcolor{oursbg} DS-Llama-8B (Ours) & \textbf{88.4} & \textbf{81.3} & 54.7 & \textbf{91.2} & \textbf{87.5} & \textbf{33.3} & \textbf{50.0} & \textbf{69.4{\tiny $\pm$0.68}} \\
\bottomrule
\end{tabular}
\end{adjustbox}
\end{threeparttable}
\caption{Accuracy (\%) on seven mathematical reasoning benchmarks with an 8K-token generation limit per call. DS-Llama-8B denotes DeepSeek-R1-Distill-Llama-8B. Avg is the unweighted average across benchmarks, with standard deviations computed over three random-seed runs.}
\label{tab:main_results_8k}
\end{table*}

\section{Experiments}

\subsection{Experimental Settings}
\label{sec:experimental_settings}

We evaluate \textsc{PIVOT} on seven mathematical reasoning benchmarks: GSM8K \citep{cobbe2021training}, Math500 \citep{hendrycks2021measuring}, Gaokao2023en \citep{zhang2023evaluating}, OlympiadBench \citep{he2024olympiadbench}, AMC23, AIME24, and AIME25. We use three base models: Qwen3-4B, Qwen3-8B \citep{qwen3technicalreport}, and DeepSeek-R1-Distill-Llama-8B \citep{deepseekai2025deepseekr1incentivizingreasoningcapability}. Performance is measured by final-answer accuracy. \textsc{PIVOT} is denoted as ``Ours'' in tables and figures.

We compare against single-sample inference and four representative inference strategies: majority voting (Majority), dynamic voting (DV) \citep{xue2023dynamic}, Best-of-N (BoN), and SCoP \citep{zhou2024paraphrase}. We impose a common upper limit of six base-model generations per instance, counting both problem solving and rewriting. Majority draws six answers and selects the most frequent answer, with ties broken at random. DV samples adaptively up to this limit, using a confidence threshold of 0.7 based on the sensitivity analysis in Table~\ref{tab:dv_threshold}. BoN generates six candidates and selects the highest-scoring answer using Skywork-Reward-V2-Llama-3.1-8B-40M \citep{liu2025skywork} as the outcome reward model, incurring additional reward-model inference cost. In our six-generation configuration, SCoP uses one paraphrase ($K=1$) and five reasoning samples ($N=5$), with the final answer selected by majority voting. PIVOT uses the default two-check configuration, requiring two, four, or six generations depending on the observed agreement, with at most six generations per instance. Thus, the compared strategies share the same maximum base-model generation budget, while DV and \textsc{PIVOT} may terminate early and use fewer generations.

We use temperature 0.6, top-$p$ 0.95, and top-$k$ 20, with three random seeds (0, 42, and 777). We set the maximum generation length per call to 8K or 16K tokens, with results reported in Tables~\ref{tab:main_results_8k} and~\ref{tab:main_results_16k}, respectively. Experiments use the Qwen2.5-Math evaluation framework\footnote{\url{https://github.com/QwenLM/Qwen2.5-Math}} and are conducted on eight NVIDIA A100 GPUs. We additionally evaluate Qwen3-32B on a subset of benchmarks to assess applicability to a larger model (Table~\ref{tab:Qwen3-32B}).

\begin{figure*}[t]
  \centering
  \includegraphics[width=\textwidth]{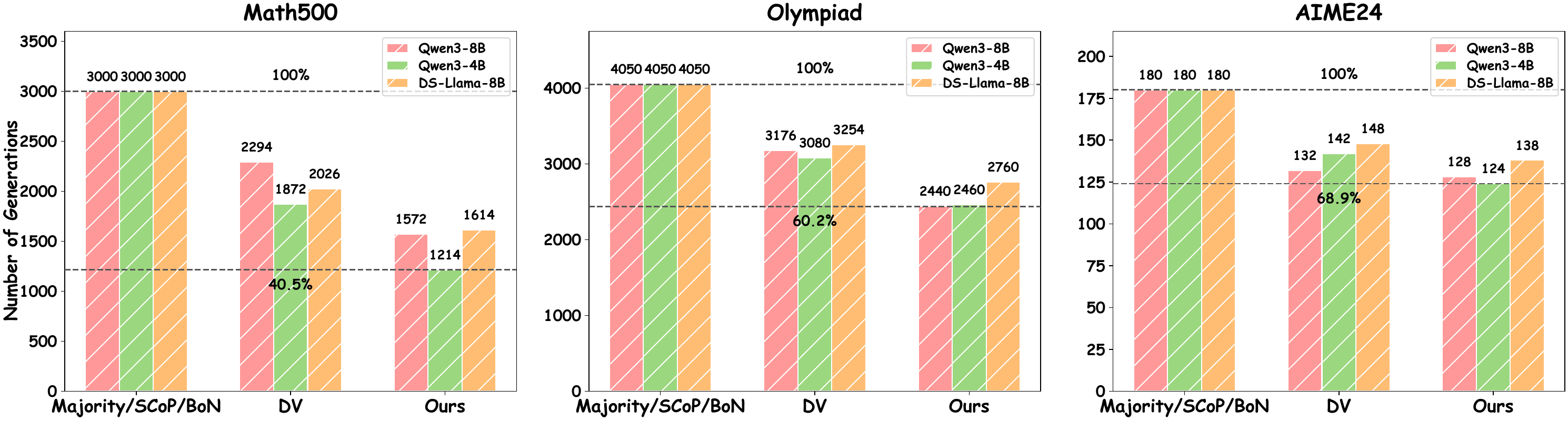}
  \caption{Total base-model generations used by different methods. Dashed lines indicate the full budget (100\%) and the minimum fraction used by our method across the three models.}
  \label{fig:sampling}
\end{figure*}

\begin{figure*}[t]
  \centering
  \includegraphics[width=\textwidth]{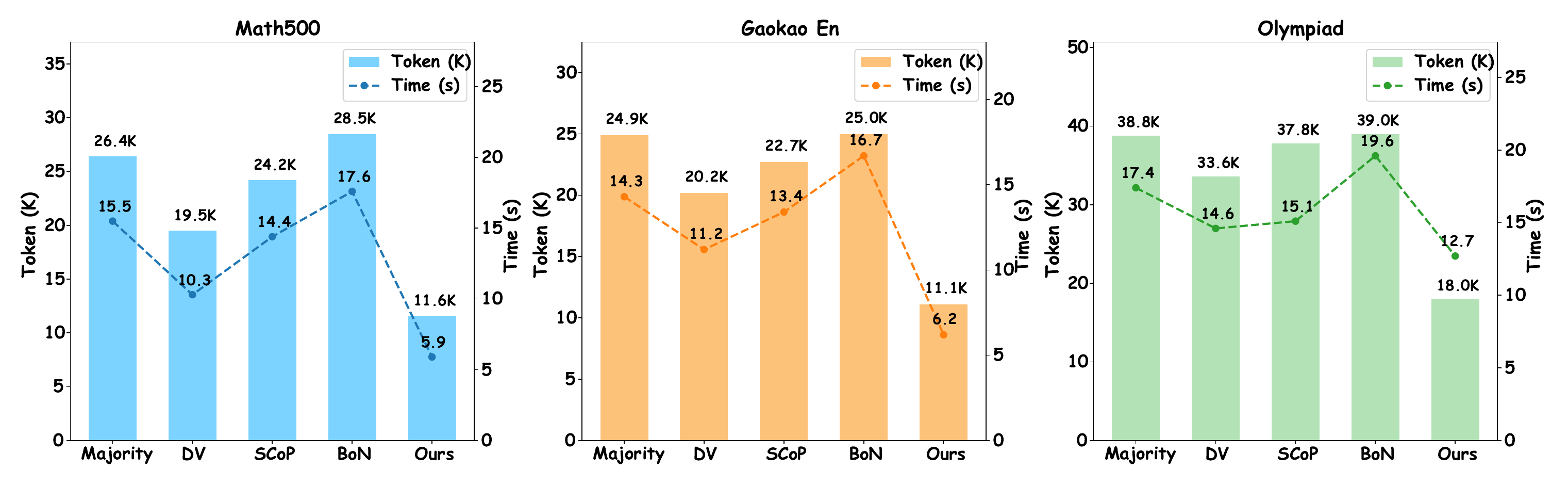}
  \caption{Average token usage and wall-clock time per instance for Qwen3-8B on Math500, Gaokao En, and Olympiad. Costs include rewriting for SCoP and our method and reward-model scoring for BoN.}
  \label{fig:token_and_time}
\end{figure*}

\subsection{Main Results}

Table~\ref{tab:main_results_8k} reports the results with an 8K-token generation limit per call. PIVOT achieves the highest average accuracy across all three models, improving over the strongest competing method for each model by 6.9, 5.2, and 2.9 percentage points on Qwen3-8B, Qwen3-4B, and DS-Llama-8B, respectively. These gains are obtained under the same maximum base-model generation budget. Results with a 16K-token generation limit in Table~\ref{tab:main_results_16k} further show that the average accuracy advantage persists when longer reasoning outputs are permitted.

On GSM8K, PIVOT yields only modest improvements over single-sample inference for Qwen3-8B and Qwen3-4B, whose baseline accuracies are already high; majority voting remains more accurate for Qwen3-4B. Table~\ref{tab:samples_8k} shows that only approximately 1.5--2\% of instances exhibit persistent disagreement after two checks for these models, so reformulation is applied to a small fraction of the benchmark. In contrast, this fraction reaches approximately 16\% for DS-Llama-8B, for which PIVOT provides a larger accuracy improvement. These results show how reformulation frequency and accuracy gains vary across models.

Figure~\ref{fig:sampling} compares the total number of base-model generations used by different methods. Across the three models, PIVOT uses 40.5--53.8\% of the maximum generation budget on Math500, 60.2--68.1\% on Olympiad, and 68.9--76.7\% on AIME24. These results show that early termination reduces generation usage while preserving the accuracy advantages reported above. Figure~\ref{fig:token_and_time} further shows that PIVOT uses fewer tokens and less wall-clock time than the compared strategies on all three evaluated benchmarks, with rewriting and reward-model scoring costs included where applicable.

\begin{table*}[t]
\centering
\small
\setlength{\tabcolsep}{5pt}
\begin{threeparttable}
\begin{adjustbox}{width=\linewidth,center}
\begin{tabular}{lccccc}
\toprule
\textbf{Model / Method} & \textbf{HumanEval} & \textbf{HumanEval+} & \textbf{MBPP} & \textbf{MBPP+} & \textbf{Avg} \\
\midrule
Qwen3-8B & 84.8 & 80.5 & 70.5 & 60.8 & 74.2\\
Qwen3-8B (Majority) & 85.5 & 79.9 & 71.0 & 60.5 & 74.2 \\
Qwen3-8B (DV) & 85.3 & 79.5 & 71.0 & 60.3 & 74.0 \\
Qwen3-8B (SCoP) & 83.7 & 79.0 & 69.9 & 60.1 & 73.2 \\
Qwen3-8B (BoN) & 85.4 & 80.7 & \textbf{72.0} & 61.4 & 74.9 \\
\rowcolor{oursbg} Qwen3-8B (Ours) & \textbf{87.8} & \textbf{81.1} & 71.7 & \textbf{61.7} & \textbf{75.6}\\
\midrule
DS-Llama-8B & 44.5 & 37.8 & 31.7 & 26.7 & 35.2 \\
DS-Llama-8B (Majority) & 48.2 & 42.9 & 32.8 & 27.1 & 37.8 \\
DS-Llama-8B (DV) & 47.9 & 41.7 & 32.8 & 26.6 & 37.3 \\
DS-Llama-8B (SCoP) & 43.7 & 38.3 & 31.2 & 25.9 & 34.8 \\
DS-Llama-8B (BoN) & 52.8 & 44.8 & 34.0 & 28.1 & 40.0 \\
\rowcolor{oursbg} DS-Llama-8B (Ours) & \textbf{54.3} & \textbf{47.0} & \textbf{38.4} & \textbf{31.2} & \textbf{42.7} \\
\midrule
Qwen2.5-Coder-7B-Instruct & 86.6 & 82.3 & 82.5 & 70.1 & 80.4 \\
Qwen2.5-Coder-7B-Instruct (Majority) & 88.1 & 82.6 & 83.9 & 69.5 & 81.0 \\
Qwen2.5-Coder-7B-Instruct (DV) & 88.1 & 82.1 & 83.6 & 69.1 & 80.7 \\
Qwen2.5-Coder-7B-Instruct (SCoP) & 87.1 & 82.2 & 81.7 & 68.4 & 79.9 \\
Qwen2.5-Coder-7B-Instruct (BoN) & 87.9 & 82.8 & 85.2 & 73.0 & 82.2 \\
\rowcolor{oursbg} Qwen2.5-Coder-7B-Instruct (Ours) & \textbf{89.0} & \textbf{83.5} & \textbf{87.8} & \textbf{73.5} & \textbf{83.5} \\
\bottomrule
\end{tabular}
\end{adjustbox}
\end{threeparttable}
\caption{Code generation accuracy (\%) on HumanEval, MBPP, and their EvalPlus variants.}
\label{tab:code}
\end{table*}

\subsection{Method Generalization Evaluation}

We evaluate \textsc{PIVOT} on code generation to examine its applicability beyond mathematical reasoning. We use the same decoding parameters and maximum base-model generation budget as in the mathematical reasoning experiments. Evaluation is conducted using EvalPlus\footnote{\url{https://github.com/evalplus/evalplus}} on HumanEval \citep{chen2021evaluatinglargelanguagemodels}, MBPP \citep{austin2021programsynthesislargelanguage}, and their variants, HumanEval+ and MBPP+. We evaluate Qwen3-8B, DeepSeek-R1-Distill-Llama-8B, and Qwen2.5-Coder-7B-Instruct \citep{hui2024qwen2}.

For code generation, we assess pairwise agreement by executing the programs on the example inputs provided in the problem statement. Two programs are treated as agreeing if their outputs match exactly on all example inputs. This criterion accommodates syntactically different implementations that exhibit the same behavior on these inputs. We use this execution-based agreement signal within the same sequential strategy-selection policy, retaining early termination upon agreement and reformulation after persistent disagreement. As shown in Table~\ref{tab:code}, \textsc{PIVOT} achieves the highest average accuracy across all three models.

% \begin{table*}[!h]
% \centering
% \small
% \begin{minipage}{0.45\linewidth}
% \centering
% \setlength{\tabcolsep}{2.5pt}
% \begin{tabular}{lccc}
% \toprule
% \textbf{Method} & \textbf{Math500} & \textbf{Olympiad} & \textbf{AIME24} \\
% \midrule
% Baseline     & 84.4 & 46.4 & 36.7 \\
% OnlyRewrite  & 90.4 & 59.0 & 56.7 \\
% OnlyMajority & 87.6 & 48.6 & 40.0 \\
% \rowcolor{oursbg} Full Method & \textbf{92.8} & \textbf{61.9} & \textbf{66.7} \\
% \bottomrule
% \end{tabular}
% \captionof{table}{Ablation accuracy (\%) on three mathematical benchmarks using Qwen3-8B.}
% \label{tab:ablation_studies}
% \end{minipage}\hfill
% \begin{minipage}{0.45\linewidth}
% \centering
% \setlength{\tabcolsep}{2.5pt}
% \begin{tabular}{lccc}
% \toprule
% \textbf{Method} & \textbf{4 generations} & \textbf{6 generations} & \textbf{8 generations} \\
% \midrule
% Majority & 87.4 & 87.6 & 87.8 \\
% DV       & 86.1 & 87.6 & 87.3 \\
% SCoP     & 85.6 & 88.2 & 87.2 \\
% BoN      & 86.2 & 85.8 & 85.8 \\
% \rowcolor{oursbg} Ours & \textbf{91.6} & \textbf{92.8} & \textbf{91.8} \\
% \bottomrule
% \end{tabular}
% \captionof{table}{Accuracy (\%) on Math500 using Qwen3-8B under maximum budgets of 4, 6, and 8 model generations.}
% \label{tab:sampling_budget}
% \end{minipage}
% \end{table*}

\section{Analysis and Discussion}

% \subsection{Effectiveness of Disagreement-Guided Routing}
\subsection{Routing Effectiveness}

We first examine whether the stability-dependent complementarity observed in Figure~\ref{fig:motivation} also holds for the partitions produced by sequential agreement checks. Table~\ref{tab:partition_strategy} provides a post-hoc comparison of majority voting and rewriting within these partitions using Qwen3-8B. Across all three benchmarks, majority voting outperforms rewriting on instances exhibiting agreement at either check, whereas rewriting outperforms majority voting on instances with persistent disagreement. Thus, the sequential checks recover a partitioning that reflects the relative utility of the two strategies, supporting their use for stability-guided strategy selection.

To assess the contribution of the routing signal, we compare \textsc{PIVOT} with a matched random-routing baseline. The baseline preserves the number of instances assigned to each route but randomly permutes assignments across instances, retaining the same route-specific operations and total generation count. In the seed-42 comparison in Table~\ref{tab:random_routing}, \textsc{PIVOT} achieves an average accuracy of 79.3\%, versus 72.6\% for random routing, a gain of 6.7 percentage points. This result supports disagreement-guided instance allocation under matched routing proportions and generation budgets. Detailed settings are provided in Appendix~\ref{app:routing_analysis}.

% \begin{table}[t]
% \centering
% \small
% \setlength{\tabcolsep}{2.5pt}
% \begin{tabular}{lccc}
% \toprule
% \textbf{Method} & \textbf{Math500} & \textbf{Olympiad} & \textbf{AIME24} \\
% \midrule
% Baseline     & 84.4 & 46.4 & 36.7 \\
% OnlyRewrite  & 90.4 & 59.0 & 56.7 \\
% OnlyMajority & 87.6 & 48.6 & 40.0 \\
% \rowcolor{oursbg}
% Full Method  & \textbf{92.8} & \textbf{61.9} & \textbf{66.7} \\
% \bottomrule
% \end{tabular}
% \caption{Ablation accuracy (\%) on three mathematical benchmarks using Qwen3-8B.}
% \label{tab:ablation_studies}
% \end{table}

% \begin{table}[t]
% \centering
% \small
% \setlength{\tabcolsep}{2.5pt}
% \begin{tabular}{lccc}
% \toprule
% \textbf{Method} & \textbf{4 generations} & \textbf{6 generations} & \textbf{8 generations} \\
% \midrule
% Majority & 87.4 & 87.6 & 87.8 \\
% DV       & 86.1 & 87.6 & 87.3 \\
% SCoP     & 85.6 & 88.2 & 87.2 \\
% BoN      & 86.2 & 85.8 & 85.8 \\
% \rowcolor{oursbg}
% Ours     & \textbf{91.6} & \textbf{92.8} & \textbf{91.8} \\
% \bottomrule
% \end{tabular}
% \caption{Accuracy (\%) on Math500 using Qwen3-8B under maximum budgets of 4, 6, and 8 model generations.}
% \label{tab:sampling_budget}
% \end{table}

\begin{figure*}[t]
  \centering
  \includegraphics[width=\textwidth]{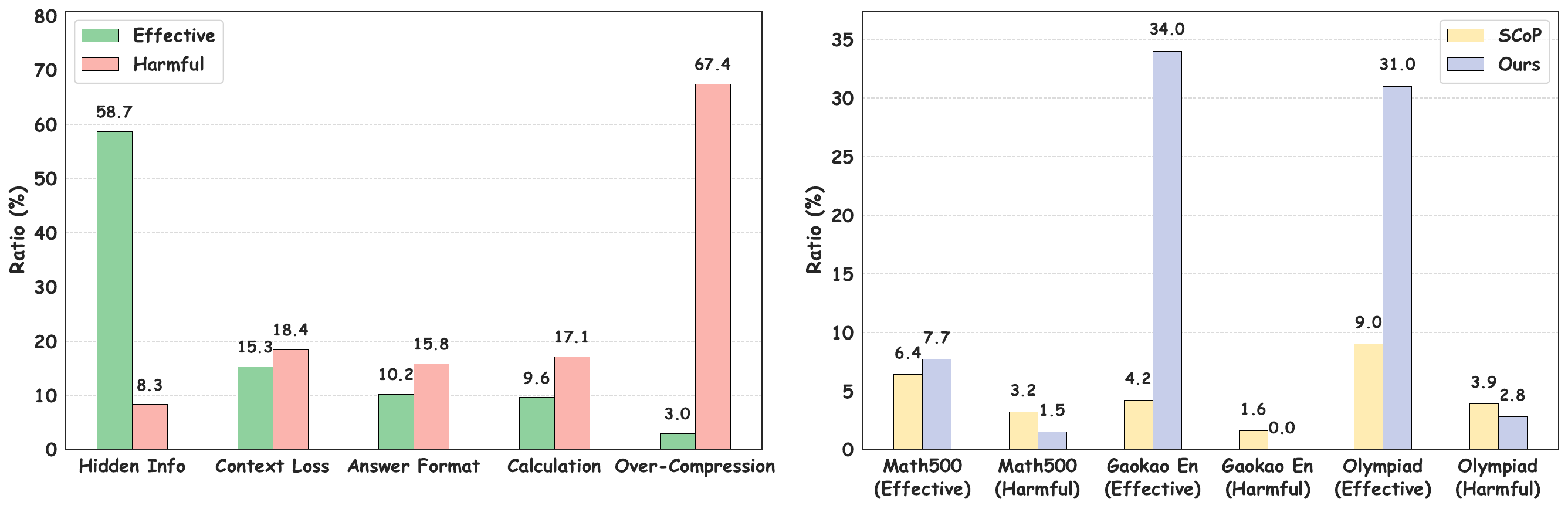}
  \caption{\textbf{Left:} Effective and harmful rewriting rates by error type. \textbf{Right:} Comparison between SCoP and our method using Qwen3-8B on Math500, Gaokao En, and Olympiad. For the right panel, rates are computed over the instances rewritten by each method.}
  \label{fig:effective_and_harmful_ratio}
\end{figure*}

\subsection{Ablation Study}

We examine the contributions of candidate aggregation and problem reformulation using Qwen3-8B on three mathematical benchmarks. Table~\ref{tab:ablation_studies} compares single-sample inference (Baseline), two restricted variants (OnlyRewrite and OnlyMajority), and the complete procedure (Full Method). Both restricted variants retain the same disagreement-based routing as the full method. OnlyRewrite allows only the rewriting branch to update provisional answers (the first sampled answer $a_1$ for each instance), whereas OnlyMajority allows only the voting branch to update them; other provisional answers remain unchanged. All variants are run independently under the same evaluation settings and averaged over the same three random seeds.

Both restricted variants improve over the baseline, with OnlyRewrite providing larger gains than OnlyMajority on all three benchmarks. Nevertheless, the complete procedure further improves over OnlyRewrite by 2.4, 2.9, and 10.0 percentage points on Math500, Olympiad, and AIME24, respectively. Since the restricted variants use the same routing rule, these results show that the improvement is not attributable solely to the rewriting branch: retaining aggregation-based answer updates also contributes to final accuracy. Together with the matched random-routing comparison, these findings support both the use of the two strategies and their disagreement-guided allocation.

We investigate how the generation budget affects performance. Experiments are conducted using Qwen3-8B on Math500, with results reported in Table~\ref{tab:sampling_budget}. Maximum budgets of four, six, and eight generations correspond to one, two, and three pairwise checks, respectively, followed by one rewrite and one solve if disagreement persists. \textsc{PIVOT} outperforms the compared methods under all three budgets, achieving accuracies of 91.6\%, 92.8\%, and 91.8\%. The default two-check configuration performs best among these settings, while increasing the number of checks does not monotonically improve accuracy. These results support the effectiveness of the policy across the tested budgets and the use of two checks as its default configuration.

\subsection{Analysis of Rewriting Applicability}

We analyze the effects of rewriting by comparing answer correctness before and after reformulation. We focus on three cases: effective rewriting, where an incorrect answer becomes correct; ineffective rewriting, where an incorrect answer remains incorrect; and harmful rewriting, where a correct answer becomes incorrect. Figure~\ref{fig:effective_and_harmful_ratio} (left) summarizes the associated error types, with representative cases provided in Appendix~\ref{sec:appendix:case_study}.

\paragraph{Effective Rewriting}
Our case analysis suggests that successful rewrites often clarify implicit conditions or mathematical relationships overlooked in the original reasoning (\textbf{hidden info error}). By reorganizing the statement and making these relationships explicit, rewriting can help the model construct a more appropriate mathematical representation and use information in the problem.

\paragraph{Ineffective Rewriting}
Rewriting may leave an answer incorrect when the failure involves output formatting (\textbf{answer format error}), loss of intermediate information (\textbf{context loss error}), or numerical computation (\textbf{calculation error}). Such failures can persist despite a clearer problem formulation.

\paragraph{Harmful Rewriting}
Excessive compression may remove or obscure necessary quantities and constraints (\textbf{over-compression error}), turning a correct answer into an incorrect one. Figure~\ref{fig:effective_and_harmful_ratio} (right) shows higher effective and lower harmful rates for \textsc{PIVOT} than SCoP. Table~\ref{tab:rewrite_prompt} reports a lower harmful rewriting rate with a phrasing-preserving prompt that does not shorten the problem.

\section{Conclusion}

We identify a stability-dependent complementarity between majority voting and problem rewriting: voting is more effective on answer-stable instances, whereas rewriting is more effective on answer-unstable instances. Building on this observation, we introduce \textsc{PIVOT}, a training-free sequential routing framework that uses persistent disagreement to guide the switch from candidate aggregation to problem reformulation. Sequential agreement checks enable early termination and reuse generated answers for both strategy selection and final answering. Experiments on seven mathematical reasoning benchmarks across three models demonstrate higher average accuracy with reduced inference cost, with additional gains on code generation. \textsc{PIVOT} also exceeds the empirical pass@6 oracle for original-formulation candidates, highlighting the value of reformulation beyond candidate aggregation. Our analysis characterizes when rewriting helps or harms, providing empirical insight into the selective use of problem reformulation.

\section*{Limitations}
\label{sec:limitations}

This work leaves room for further exploration in certain aspects.
\begin{itemize}[left=0pt]
\item The routing signal reflects answer stability under the current problem formulation. Stable but incorrect answers may also satisfy the agreement checks, motivating the integration of complementary correctness-sensitive signals.
\item Our case analysis identifies the omission of necessary constraints as a source of harmful rewriting. Improving semantic preservation during reformulation could further strengthen the reliability of the approach.
\item The framework examines the allocation of computation between candidate aggregation and problem reformulation. Extending stability-guided selection to strategies such as search and verification offers a direction for broader investigation.
\end{itemize}

% \section*{Acknowledgments}

% This document has been adapted
% by Steven Bethard, Ryan Cotterell and Rui Yan
% from the instructions for earlier ACL and NAACL proceedings, including those for
% ACL 2019 by Douwe Kiela and Ivan Vuli\'{c},
% NAACL 2019 by Stephanie Lukin and Alla Roskovskaya,
% ACL 2018 by Shay Cohen, Kevin Gimpel, and Wei Lu,
% NAACL 2018 by Margaret Mitchell and Stephanie Lukin,
% Bib\TeX{} suggestions for (NA)ACL 2017/2018 from Jason Eisner,
% ACL 2017 by Dan Gildea and Min-Yen Kan,
% NAACL 2017 by Margaret Mitchell,
% ACL 2012 by Maggie Li and Michael White,
% ACL 2010 by Jing-Shin Chang and Philipp Koehn,
% ACL 2008 by Johanna D. Moore, Simone Teufel, James Allan, and Sadaoki Furui,
% ACL 2005 by Hwee Tou Ng and Kemal Oflazer,
% ACL 2002 by Eugene Charniak and Dekang Lin,
% and earlier ACL and EACL formats written by several people, including
% John Chen, Henry S. Thompson and Donald Walker.
% Additional elements were taken from the formatting instructions of the \emph{International Joint Conference on Artificial Intelligence} and the \emph{Conference on Computer Vision and Pattern Recognition}.


\bibliography{custom}

\appendix
\clearpage
% Thematic appendix layout (v5).
% Packages: booktabs, array, tabularx, graphicx, caption, xcolor[table], tcolorbox[skins].
% Original labels and Table 3--15 / Figure 7 numbering are retained.
\clearpage
\appendix
\begingroup
\raggedbottom
\newsavebox{\appmeasurebox}
\newsavebox{\appleftbox}
\newsavebox{\apprightbox}
\newlength{\apppairheight}
\twocolumn[{
\section{Experimental Results}
\label{sec:experimental_results}
\noindent\begin{minipage}[t]{\columnwidth}\vspace{0pt}
Table~\ref{tab:main_results_16k} extends the main comparison to a 16K-token generation limit per call. It reports the same seven mathematical benchmarks and three models, allowing the comparison to be examined when longer reasoning outputs are permitted.
\end{minipage}\hfill
\begin{minipage}[t]{\columnwidth}\vspace{0pt}
Tables~\ref{tab:easy_recall}--\ref{tab:sampling_budget} examine agreement thresholds, component contributions, and generation budgets. They complement the main accuracy results by separating the effects of the agreement criterion, the retained inference branches, and the maximum number of generations.
\end{minipage}\par
\par\vspace{16pt}
\begingroup
\small
\setlength{\tabcolsep}{2.5pt}
\renewcommand{\arraystretch}{1.08}
\captionsetup{skip=10pt}
\centering
\sbox{\appmeasurebox}{\begin{tabular}{lcccccccc}
\toprule
\textbf{Model / Method} & \textbf{Math500} & \textbf{Gaokao En} & \textbf{Olympiad} & \textbf{GSM8K} & \textbf{AMC23} & \textbf{AIME25} & \textbf{AIME24} & \textbf{Avg} \\
\midrule
Qwen3-8B & 91.6 & 82.9 & 60.7 & 94.9 & 90.0 & 46.7 & 66.7 & 76.2 \\
Qwen3-8B (Majority) & 93.9 & 84.2 & 64.7 & \textbf{96.4} & \textbf{95.0} & 53.3 & 70.0 & 79.6 \\
Qwen3-8B (DV) & 92.9 & 84.4 & 64.5 & 95.9 & 92.5 & 50.0 & 70.0 & 78.6 \\
Qwen3-8B (SCoP) & 92.4 & 84.5 & 65.6 & 95.2 & \textbf{95.0} & 53.3 & 63.3 & 78.5 \\
Qwen3-8B (BoN) & 93.4 & 83.6 & 64.6 & 95.7 & 90.0 & 53.3 & 63.3 & 77.7 \\
\rowcolor{oursbg} Qwen3-8B (Ours) & \textbf{95.2} & \textbf{86.0} & \textbf{67.3} & 96.1 & \textbf{95.0} & \textbf{66.7} & \textbf{80.0} & \textbf{83.8} \\
\midrule
Qwen3-4B & 91.8 & 83.6 & 63.4 & 95.0 & 92.5 & 46.7 & 66.7 & 77.1 \\
Qwen3-4B (Majority) & 95.2 & 84.9 & 66.5 & 95.0 & 95.0 & 60.0 & \textbf{70.0} & 81.0 \\
Qwen3-4B (DV) & 95.0 & 84.7 & 66.3 & 94.9 & 92.5 & 55.0 & 66.7 & 79.3 \\
Qwen3-4B (SCoP) & 92.0 & 84.7 & 64.6 & 93.6 & 92.5 & 46.7 & 60.0 & 76.3 \\
Qwen3-4B (BoN) & 92.8 & 82.1 & 64.1 & 94.8 & 90.0 & 60.0 & 66.7 & 78.6 \\
\rowcolor{oursbg} Qwen3-4B (Ours) & \textbf{95.6} & \textbf{85.7} & \textbf{67.1} & \textbf{95.1} & \textbf{97.5} & \textbf{70.0} & \textbf{70.0} & \textbf{83.0} \\
\midrule
DS-Llama-8B & 82.4 & 73.8 & 54.1 & 85.0 & 87.5 & 33.3 & 40.0 & 65.2 \\
DS-Llama-8B (Majority) & 88.8 & 80.8 & 58.1 & 89.0 & 87.5 & 40.0 & 56.7 & 71.6 \\
DS-Llama-8B (DV) & 88.8 & 80.8 & 57.9 & 89.0 & 85.0 & 36.7 & 53.3 & 70.2 \\
DS-Llama-8B (SCoP) & 88.4 & 78.2 & 55.3 & 89.3 & 87.5 & 33.3 & 43.3 & 67.9 \\
DS-Llama-8B (BoN) & 87.2 & 79.4 & 57.1 & 90.5 & 85.0 & 36.7 & 50.0 & 69.4 \\
\rowcolor{oursbg} DS-Llama-8B (Ours) & \textbf{91.0} & \textbf{81.2} & \textbf{58.5} & \textbf{90.8} & \textbf{92.5} & \textbf{43.3} & \textbf{60.0} & \textbf{73.9} \\
\bottomrule
\end{tabular}}
\setlength{\tabcolsep}{\dimexpr\tabcolsep+(\linewidth-\wd\appmeasurebox)/18\relax}
\begin{tabular}{lcccccccc}
\toprule
\textbf{Model / Method} & \textbf{Math500} & \textbf{Gaokao En} & \textbf{Olympiad} & \textbf{GSM8K} & \textbf{AMC23} & \textbf{AIME25} & \textbf{AIME24} & \textbf{Avg} \\
\midrule
Qwen3-8B & 91.6 & 82.9 & 60.7 & 94.9 & 90.0 & 46.7 & 66.7 & 76.2 \\
Qwen3-8B (Majority) & 93.9 & 84.2 & 64.7 & \textbf{96.4} & \textbf{95.0} & 53.3 & 70.0 & 79.6 \\
Qwen3-8B (DV) & 92.9 & 84.4 & 64.5 & 95.9 & 92.5 & 50.0 & 70.0 & 78.6 \\
Qwen3-8B (SCoP) & 92.4 & 84.5 & 65.6 & 95.2 & \textbf{95.0} & 53.3 & 63.3 & 78.5 \\
Qwen3-8B (BoN) & 93.4 & 83.6 & 64.6 & 95.7 & 90.0 & 53.3 & 63.3 & 77.7 \\
\rowcolor{oursbg} Qwen3-8B (Ours) & \textbf{95.2} & \textbf{86.0} & \textbf{67.3} & 96.1 & \textbf{95.0} & \textbf{66.7} & \textbf{80.0} & \textbf{83.8} \\
\midrule
Qwen3-4B & 91.8 & 83.6 & 63.4 & 95.0 & 92.5 & 46.7 & 66.7 & 77.1 \\
Qwen3-4B (Majority) & 95.2 & 84.9 & 66.5 & 95.0 & 95.0 & 60.0 & \textbf{70.0} & 81.0 \\
Qwen3-4B (DV) & 95.0 & 84.7 & 66.3 & 94.9 & 92.5 & 55.0 & 66.7 & 79.3 \\
Qwen3-4B (SCoP) & 92.0 & 84.7 & 64.6 & 93.6 & 92.5 & 46.7 & 60.0 & 76.3 \\
Qwen3-4B (BoN) & 92.8 & 82.1 & 64.1 & 94.8 & 90.0 & 60.0 & 66.7 & 78.6 \\
\rowcolor{oursbg} Qwen3-4B (Ours) & \textbf{95.6} & \textbf{85.7} & \textbf{67.1} & \textbf{95.1} & \textbf{97.5} & \textbf{70.0} & \textbf{70.0} & \textbf{83.0} \\
\midrule
DS-Llama-8B & 82.4 & 73.8 & 54.1 & 85.0 & 87.5 & 33.3 & 40.0 & 65.2 \\
DS-Llama-8B (Majority) & 88.8 & 80.8 & 58.1 & 89.0 & 87.5 & 40.0 & 56.7 & 71.6 \\
DS-Llama-8B (DV) & 88.8 & 80.8 & 57.9 & 89.0 & 85.0 & 36.7 & 53.3 & 70.2 \\
DS-Llama-8B (SCoP) & 88.4 & 78.2 & 55.3 & 89.3 & 87.5 & 33.3 & 43.3 & 67.9 \\
DS-Llama-8B (BoN) & 87.2 & 79.4 & 57.1 & 90.5 & 85.0 & 36.7 & 50.0 & 69.4 \\
\rowcolor{oursbg} DS-Llama-8B (Ours) & \textbf{91.0} & \textbf{81.2} & \textbf{58.5} & \textbf{90.8} & \textbf{92.5} & \textbf{43.3} & \textbf{60.0} & \textbf{73.9} \\
\bottomrule
\end{tabular}
\par
\captionof{table}{Accuracy (\%) on seven mathematical reasoning benchmarks with a 16K-token generation limit per call.}
\label{tab:main_results_16k}
\endgroup
\par\vspace{10pt}
\par\vspace{10pt}
\sbox{\appleftbox}{\begin{minipage}[t]{0.485\textwidth}\vspace{0pt}
\begingroup
\small
\setlength{\tabcolsep}{2.5pt}
\renewcommand{\arraystretch}{1.2125}
\captionsetup{skip=10pt}
\centering
\sbox{\appmeasurebox}{\begin{tabular}{lcccc}
\toprule
\textbf{Threshold $n$} & \textbf{Math500} & \textbf{Gaokao En} & \textbf{Olympiad} & \textbf{GSM8K} \\
\midrule
n=2 & 94.5 & 96.2 & 87.5 & 99.0 \\
n=3 & 91.4 & 94.9 & 83.2 & 97.2 \\
n=4 & 89.1 & 89.6 & 77.9 & 97.1 \\
n=5 & 84.5 & 87.2 & 72.4 & 96.9 \\
n=6 & 82.7 & 85.6 & 69.5 & 96.1 \\
\bottomrule
\end{tabular}}
\setlength{\tabcolsep}{\dimexpr\tabcolsep+(\linewidth-\wd\appmeasurebox)/10\relax}
\begin{tabular}{lcccc}
\toprule
\textbf{Threshold $n$} & \textbf{Math500} & \textbf{Gaokao En} & \textbf{Olympiad} & \textbf{GSM8K} \\
\midrule
n=2 & 94.5 & 96.2 & 87.5 & 99.0 \\
n=3 & 91.4 & 94.9 & 83.2 & 97.2 \\
n=4 & 89.1 & 89.6 & 77.9 & 97.1 \\
n=5 & 84.5 & 87.2 & 72.4 & 96.9 \\
n=6 & 82.7 & 85.6 & 69.5 & 96.1 \\
\bottomrule
\end{tabular}
\par
\captionof{table}{Recall (\%) of correct instances when at least $n$ sampled answers agree.}
\label{tab:easy_recall}
\endgroup
\par\vspace{10pt}
\end{minipage}}
\sbox{\apprightbox}{\begin{minipage}[t]{0.485\textwidth}\vspace{0pt}
\begingroup
\small
\setlength{\tabcolsep}{2.5pt}
\renewcommand{\arraystretch}{1.455}
\captionsetup{skip=10pt}
\centering
\sbox{\appmeasurebox}{\begin{tabular}{lccc}
\toprule
\textbf{Method} & \textbf{Math500} & \textbf{Olympiad} & \textbf{AIME24} \\
\midrule
Baseline     & 84.4 & 46.4 & 36.7 \\
OnlyRewrite  & 90.4 & 59.0 & 56.7 \\
OnlyMajority & 87.6 & 48.6 & 40.0 \\
\rowcolor{oursbg}
Full Method  & \textbf{92.8} & \textbf{61.9} & \textbf{66.7} \\
\bottomrule
\end{tabular}}
\setlength{\tabcolsep}{\dimexpr\tabcolsep+(\linewidth-\wd\appmeasurebox)/8\relax}
\begin{tabular}{lccc}
\toprule
\textbf{Method} & \textbf{Math500} & \textbf{Olympiad} & \textbf{AIME24} \\
\midrule
Baseline     & 84.4 & 46.4 & 36.7 \\
OnlyRewrite  & 90.4 & 59.0 & 56.7 \\
OnlyMajority & 87.6 & 48.6 & 40.0 \\
\rowcolor{oursbg}
Full Method  & \textbf{92.8} & \textbf{61.9} & \textbf{66.7} \\
\bottomrule
\end{tabular}
\par
\captionof{table}{Ablation accuracy (\%) on three mathematical benchmarks using Qwen3-8B.}
\label{tab:ablation_studies}
\endgroup
\par\vspace{10pt}
\end{minipage}}
\setlength{\apppairheight}{\dimexpr\ht\appleftbox+\dp\appleftbox\relax}
\ifdim\dimexpr\ht\apprightbox+\dp\apprightbox\relax>\apppairheight
\setlength{\apppairheight}{\dimexpr\ht\apprightbox+\dp\apprightbox\relax}\fi
\noindent\begin{minipage}[t][\apppairheight][t]{0.485\textwidth}\vspace{0pt}\usebox{\appleftbox}\end{minipage}\hfill
\begin{minipage}[t][\apppairheight][t]{0.485\textwidth}\vspace{0pt}\usebox{\apprightbox}\end{minipage}\par
\par\vspace{10pt}
\begingroup
\small
\setlength{\tabcolsep}{2.5pt}
\renewcommand{\arraystretch}{1.08}
\captionsetup{skip=10pt}
\centering
\begin{minipage}{0.78\textwidth}
\sbox{\appmeasurebox}{\begin{tabular}{lccc}
\toprule
\textbf{Method} & \textbf{4 generations} & \textbf{6 generations} & \textbf{8 generations} \\
\midrule
Majority & 87.4 & 87.6 & 87.8 \\
DV       & 86.1 & 87.6 & 87.3 \\
SCoP     & 85.6 & 88.2 & 87.2 \\
BoN      & 86.2 & 85.8 & 85.8 \\
\rowcolor{oursbg}
Ours     & \textbf{91.6} & \textbf{92.8} & \textbf{91.8} \\
\bottomrule
\end{tabular}}
\setlength{\tabcolsep}{\dimexpr\tabcolsep+(\linewidth-\wd\appmeasurebox)/8\relax}
\begin{tabular}{lccc}
\toprule
\textbf{Method} & \textbf{4 generations} & \textbf{6 generations} & \textbf{8 generations} \\
\midrule
Majority & 87.4 & 87.6 & 87.8 \\
DV       & 86.1 & 87.6 & 87.3 \\
SCoP     & 85.6 & 88.2 & 87.2 \\
BoN      & 86.2 & 85.8 & 85.8 \\
\rowcolor{oursbg}
Ours     & \textbf{91.6} & \textbf{92.8} & \textbf{91.8} \\
\bottomrule
\end{tabular}
\end{minipage}
\par
\captionof{table}{Accuracy (\%) on Math500 using Qwen3-8B under maximum budgets of 4, 6, and 8 model generations.}
\label{tab:sampling_budget}
\endgroup
\par\vspace{10pt}
}]
\clearpage
\twocolumn[{
\subsection*{Diagnostic Partitions and Strategy Comparison}
\noindent\begin{minipage}[t]{\columnwidth}\vspace{0pt}
\paragraph{Motivation analysis.}
For Figure~\ref{fig:motivation}, we first partition instances using six independently sampled answers. Base, Majority, and Rewriting are then evaluated independently within each fixed subset using fresh generations. These stability-defined subsets are distinct from the sequential routing partitions evaluated below.
\end{minipage}\hfill
\begin{minipage}[t]{\columnwidth}\vspace{0pt}
\paragraph{Diagnostic protocol.}
For Tables~\ref{tab:samples_8k}, \ref{tab:samples_correct} and \ref{tab:partition_strategy}, we independently rerun the disagreement-based procedure to obtain a fixed Check 1 / Check 2 / Persistent partition. We then evaluate Baseline, six-sample Majority, and Rewrite on each partition using newly generated outputs. These diagnostic experiments use the same benchmark instances, model, and decoding settings as the main evaluation.

\end{minipage}\par
\par\vspace{12pt}
\begingroup
\small
\setlength{\tabcolsep}{2.5pt}
\renewcommand{\arraystretch}{1.08}
\captionsetup{skip=10pt}
\centering
\sbox{\appmeasurebox}{\begin{tabular}{lccccccc}
\toprule
\textbf{Model / Samples} & \textbf{Math500} & \textbf{Gaokao En} & \textbf{Olympiad} & \textbf{GSM8K} & \textbf{AMC23} & \textbf{AIME25} & \textbf{AIME24} \\
\midrule
Qwen3-8B (Check 1) & 66.4 & 62.9 & 55.3 & 97.6 & 65.0 & 26.7 & 40.0 \\
Qwen3-8B (Check 2) & 10.0 & 10.1 & 8.7 & 0.9 & 10.0 & 13.3 & 6.7  \\
Qwen3-8B (Persistent) & 23.6 & 27.0 & 36.0 & 1.5 & 25.0 & 60.0 & 53.3 \\
\midrule
Qwen3-4B (Check 1) & 87.4 & 81.3 & 55.3 & 96.0 & 67.5 & 26.7 & 43.3 \\
Qwen3-4B (Check 2) & 3.8 & 7.0 & 7.3 & 2.0 & 2.5 & 6.7 & 6.7 \\
Qwen3-4B (Persistent) & 8.8 & 11.7 & 37.4 & 2.0 & 30.0 & 66.6 & 50.0 \\
\midrule
DS-Llama-8B (Check 1) & 63.2 & 54.3 & 43.1 & 69.6 & 70.0 & 20.0 & 26.7 \\
DS-Llama-8B (Check 2) & 12.2 & 17.1 & 13.5 & 14.3 & 10.0 & 0.0 & 16.7 \\
DS-Llama-8B (Persistent) & 24.6 & 28.6 & 43.4 & 16.1 & 20.0 & 80.0 & 56.6 \\
\bottomrule
\end{tabular}}
\setlength{\tabcolsep}{\dimexpr\tabcolsep+(\linewidth-\wd\appmeasurebox)/16\relax}
\begin{tabular}{lccccccc}
\toprule
\textbf{Model / Samples} & \textbf{Math500} & \textbf{Gaokao En} & \textbf{Olympiad} & \textbf{GSM8K} & \textbf{AMC23} & \textbf{AIME25} & \textbf{AIME24} \\
\midrule
Qwen3-8B (Check 1) & 66.4 & 62.9 & 55.3 & 97.6 & 65.0 & 26.7 & 40.0 \\
Qwen3-8B (Check 2) & 10.0 & 10.1 & 8.7 & 0.9 & 10.0 & 13.3 & 6.7  \\
Qwen3-8B (Persistent) & 23.6 & 27.0 & 36.0 & 1.5 & 25.0 & 60.0 & 53.3 \\
\midrule
Qwen3-4B (Check 1) & 87.4 & 81.3 & 55.3 & 96.0 & 67.5 & 26.7 & 43.3 \\
Qwen3-4B (Check 2) & 3.8 & 7.0 & 7.3 & 2.0 & 2.5 & 6.7 & 6.7 \\
Qwen3-4B (Persistent) & 8.8 & 11.7 & 37.4 & 2.0 & 30.0 & 66.6 & 50.0 \\
\midrule
DS-Llama-8B (Check 1) & 63.2 & 54.3 & 43.1 & 69.6 & 70.0 & 20.0 & 26.7 \\
DS-Llama-8B (Check 2) & 12.2 & 17.1 & 13.5 & 14.3 & 10.0 & 0.0 & 16.7 \\
DS-Llama-8B (Persistent) & 24.6 & 28.6 & 43.4 & 16.1 & 20.0 & 80.0 & 56.6 \\
\bottomrule
\end{tabular}
\par
\captionof{table}{Distribution of instances (\%) across the three routing partitions, with an 8K-token generation limit per call. Check 1 and Check 2 indicate the first agreeing pair; Persistent indicates disagreement across both checks.}
\label{tab:samples_8k}
\endgroup
\par\vspace{10pt}
\par\vspace{8pt}
\begingroup
\small
\setlength{\tabcolsep}{2.5pt}
\renewcommand{\arraystretch}{1.08}
\captionsetup{skip=10pt}
\centering
\sbox{\appmeasurebox}{\begin{tabular}{lcccc}
\toprule
\textbf{Datasets} & \textbf{Check 1} & \textbf{Check 2} & \textbf{Persistent} & \textbf{Drop (pp)} \\
\midrule
Olympiad & 63.4 & 11.1 & 4.1 & $\downarrow$ 59.3 \\
Gaokao En & 87.3 & 52.6 & 20.0 & $\downarrow$ 67.3 \\
Math500 & 94.2 & 65.2 & 33.3 & $\downarrow$ 60.9 \\
\bottomrule
\end{tabular}}
\setlength{\tabcolsep}{\dimexpr\tabcolsep+(\linewidth-\wd\appmeasurebox)/10\relax}
\begin{tabular}{lcccc}
\toprule
\textbf{Datasets} & \textbf{Check 1} & \textbf{Check 2} & \textbf{Persistent} & \textbf{Drop (pp)} \\
\midrule
Olympiad & 63.4 & 11.1 & 4.1 & $\downarrow$ 59.3 \\
Gaokao En & 87.3 & 52.6 & 20.0 & $\downarrow$ 67.3 \\
Math500 & 94.2 & 65.2 & 33.3 & $\downarrow$ 60.9 \\
\bottomrule
\end{tabular}
\par
\captionof{table}{Conditional accuracy (\%) of Baseline within the routing partitions using Qwen3-8B. Drop is the accuracy difference between Check 1 and Persistent.}
\label{tab:samples_correct}
\endgroup
\par\vspace{10pt}
\par\vspace{8pt}
\begingroup
\small
\setlength{\tabcolsep}{2.5pt}
\renewcommand{\arraystretch}{1.08}
\captionsetup{skip=10pt}
\centering
\sbox{\appmeasurebox}{\begin{tabular}{lccccccccc}
\toprule
\textbf{Method} 
& \multicolumn{3}{c}{\textbf{Olympiad}} 
& \multicolumn{3}{c}{\textbf{Gaokao En}} 
& \multicolumn{3}{c}{\textbf{Math500}} \\
\cmidrule(lr){2-4} \cmidrule(lr){5-7} \cmidrule(lr){8-10}
& \textbf{Check 1} & \textbf{Check 2} & \textbf{Persistent}
& \textbf{Check 1} & \textbf{Check 2} & \textbf{Persistent}
& \textbf{Check 1} & \textbf{Check 2} & \textbf{Persistent} \\
\midrule
Baseline 
& 63.4 & 11.1 & 4.1
& 87.3 & 52.6 & 20.0
& 94.2 & 65.2 & 33.3 \\

Majority 
& 65.4 & 44.8 & 4.1
& 88.0 & 63.2 & 28.0
& 96.1 & 68.0 & 39.7  \\

Rewrite 
& 61.1 & 38.1 & 27.8 
& 85.8 & 55.2 & 58.0 
& 93.4 & 66.5 & 68.3  \\

\bottomrule
\end{tabular}}
\setlength{\tabcolsep}{\dimexpr\tabcolsep+(\linewidth-\wd\appmeasurebox)/20\relax}
\begin{tabular}{lccccccccc}
\toprule
\textbf{Method} 
& \multicolumn{3}{c}{\textbf{Olympiad}} 
& \multicolumn{3}{c}{\textbf{Gaokao En}} 
& \multicolumn{3}{c}{\textbf{Math500}} \\
\cmidrule(lr){2-4} \cmidrule(lr){5-7} \cmidrule(lr){8-10}
& \textbf{Check 1} & \textbf{Check 2} & \textbf{Persistent}
& \textbf{Check 1} & \textbf{Check 2} & \textbf{Persistent}
& \textbf{Check 1} & \textbf{Check 2} & \textbf{Persistent} \\
\midrule
Baseline 
& 63.4 & 11.1 & 4.1
& 87.3 & 52.6 & 20.0
& 94.2 & 65.2 & 33.3 \\

Majority 
& 65.4 & 44.8 & 4.1
& 88.0 & 63.2 & 28.0
& 96.1 & 68.0 & 39.7  \\

Rewrite 
& 61.1 & 38.1 & 27.8 
& 85.8 & 55.2 & 58.0 
& 93.4 & 66.5 & 68.3  \\

\bottomrule
\end{tabular}
\par
\captionof{table}{Post-hoc strategy accuracy (\%) across routing partitions on Olympiad, Gaokao En, and Math500 using Qwen3-8B. Each partition is evaluated separately with six-sample majority voting and with one rewrite followed by a single solve. Check 1 and Check 2 denote instances first exhibiting agreement at the respective check; Persistent denotes disagreement across both checks.}
\label{tab:partition_strategy}
\endgroup
\par\vspace{10pt}
\noindent\begin{minipage}[t]{\columnwidth}\vspace{0pt}
\paragraph{Reading the partitions.} Check 1 and Check 2 identify the first check at which an agreeing pair is observed. Persistent identifies instances for which both checks disagree. These labels describe the observed check sequence rather than answer correctness.
\end{minipage}\hfill
\begin{minipage}[t]{\columnwidth}\vspace{0pt}
Table~\ref{tab:partition_strategy} compares the two strategies within the same diagnostic partitions. The end-to-end route counts and final accuracies are reported separately in Table~\ref{tab:route_statistics}, distinguishing conditional strategy comparisons from evaluation of the complete policy.
\end{minipage}\par

}]
\clearpage
\twocolumn[{
\subsection*{Additional Evaluation and Filtering Analysis}
\noindent\begin{minipage}[t]{\columnwidth}\vspace{0pt}
\paragraph{Dynamic-voting threshold.} Table~\ref{tab:dv_threshold} reports DV accuracy for thresholds from 0.5 to 0.9. The selected threshold, 0.7, achieves the highest or tied-highest accuracy on each of the three reported benchmarks.
\end{minipage}\hfill
\begin{minipage}[t]{\columnwidth}\vspace{0pt}
\paragraph{Larger-model evaluation.} Table~\ref{tab:Qwen3-32B} reports the comparison using Qwen3-32B. Our method achieves the highest accuracy on Math500, Olympiad, and Gaokao En, extending the evaluation beyond the smaller models in the main table.
\end{minipage}\par
\par\vspace{12pt}
\sbox{\appleftbox}{\begin{minipage}[t]{0.485\textwidth}\vspace{0pt}
\begingroup
\small
\setlength{\tabcolsep}{2.5pt}
\renewcommand{\arraystretch}{1.26}
\captionsetup{skip=10pt}
\centering
\sbox{\appmeasurebox}{\begin{tabular}{lccc}
\toprule
\textbf{Threshold} & \textbf{Math500} & \textbf{Olympiad} & \textbf{Gaokao En} \\
\midrule
0.5 & 84.0 & 44.9 & 75.1 \\
0.6 & 87.4 & 49.2 & 79.2 \\
\rowcolor{oursbg} 0.7 & 87.6 & 49.5 & 79.7 \\
0.8 & 87.6 & 49.2 & 79.7 \\
0.9 & 87.6 & 49.0 & 79.5 \\
\bottomrule
\end{tabular}}
\setlength{\tabcolsep}{\dimexpr\tabcolsep+(\linewidth-\wd\appmeasurebox)/8\relax}
\begin{tabular}{lccc}
\toprule
\textbf{Threshold} & \textbf{Math500} & \textbf{Olympiad} & \textbf{Gaokao En} \\
\midrule
0.5 & 84.0 & 44.9 & 75.1 \\
0.6 & 87.4 & 49.2 & 79.2 \\
\rowcolor{oursbg} 0.7 & 87.6 & 49.5 & 79.7 \\
0.8 & 87.6 & 49.2 & 79.7 \\
0.9 & 87.6 & 49.0 & 79.5 \\
\bottomrule
\end{tabular}
\par
\captionof{table}{Accuracy (\%) of dynamic voting (DV) under different thresholds on three benchmarks.}
\label{tab:dv_threshold}
\endgroup
\par\vspace{10pt}
\end{minipage}}
\sbox{\apprightbox}{\begin{minipage}[t]{0.485\textwidth}\vspace{0pt}
\begingroup
\small
\setlength{\tabcolsep}{2.5pt}
\renewcommand{\arraystretch}{1.08}
\captionsetup{skip=10pt}
\centering
\sbox{\appmeasurebox}{\begin{tabular}{lccc}
\toprule
\textbf{Method} &\textbf{Math500} & \textbf{Olympiad} & \textbf{Gaokao En} \\
\midrule
Baseline & 87.8 & 50.5 & 79.5 \\
Majority & 88.2 & 54.3 & 80.7 \\
DV       & 88.2 & 54.0 & 80.0 \\ 
SCoP     & 89.5 & 57.7 & 82.3 \\
BoN      & 88.6 & 55.6 & 79.7 \\
\rowcolor{oursbg} Ours & \textbf{93.1} & \textbf{62.8} & \textbf{84.2} \\
\bottomrule
\end{tabular}}
\setlength{\tabcolsep}{\dimexpr\tabcolsep+(\linewidth-\wd\appmeasurebox)/8\relax}
\begin{tabular}{lccc}
\toprule
\textbf{Method} &\textbf{Math500} & \textbf{Olympiad} & \textbf{Gaokao En} \\
\midrule
Baseline & 87.8 & 50.5 & 79.5 \\
Majority & 88.2 & 54.3 & 80.7 \\
DV       & 88.2 & 54.0 & 80.0 \\ 
SCoP     & 89.5 & 57.7 & 82.3 \\
BoN      & 88.6 & 55.6 & 79.7 \\
\rowcolor{oursbg} Ours & \textbf{93.1} & \textbf{62.8} & \textbf{84.2} \\
\bottomrule
\end{tabular}
\par
\captionof{table}{Comparison of different methods on three benchmarks using Qwen3-32B.}
\label{tab:Qwen3-32B}
\endgroup
\par\vspace{10pt}
\end{minipage}}
\setlength{\apppairheight}{\dimexpr\ht\appleftbox+\dp\appleftbox\relax}
\ifdim\dimexpr\ht\apprightbox+\dp\apprightbox\relax>\apppairheight
\setlength{\apppairheight}{\dimexpr\ht\apprightbox+\dp\apprightbox\relax}\fi
\noindent\begin{minipage}[t][\apppairheight][t]{0.485\textwidth}\vspace{0pt}\usebox{\appleftbox}\end{minipage}\hfill
\begin{minipage}[t][\apppairheight][t]{0.485\textwidth}\vspace{0pt}\usebox{\apprightbox}\end{minipage}\par
\par\vspace{12pt}
\noindent\begin{minipage}[t]{\columnwidth}\vspace{0pt}
\paragraph{Repeated filtering.} Figure~\ref{fig:recall} reports the recall of incorrect instances together with cumulative sampling cost as filtering continues. This analysis displays both quantities along the same iteration axis.
\end{minipage}\hfill
\begin{minipage}[t]{\columnwidth}\vspace{0pt}
Across the three benchmarks, longer filtering reduces recall while cumulative sampling cost increases. Table~\ref{tab:sampling_budget} separately evaluates the complete policy under different maximum generation budgets, connecting the diagnostic-cost analysis with end-to-end accuracy.
\end{minipage}\par
\par\vspace{10pt}
\centering
\includegraphics[width=\textwidth]{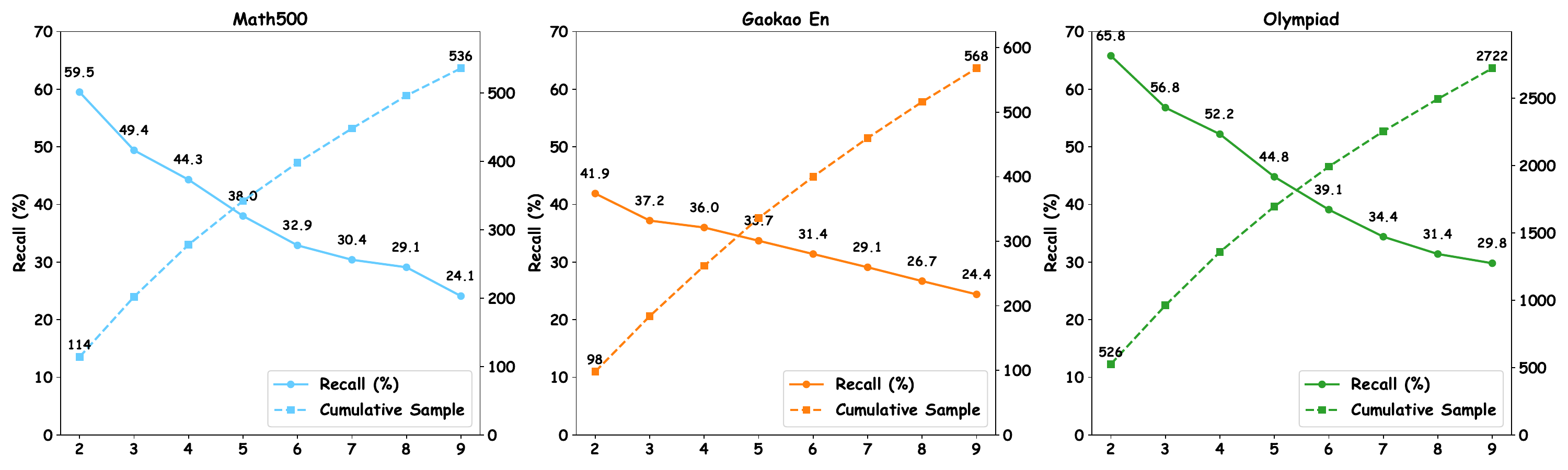}
\par
\captionsetup{skip=10pt}
\captionof{figure}{Recall (\%) of incorrect instances and cumulative sampling cost as the number of filtering iterations increases on Math500, Gaokao En, and Olympiad. Solid lines show recall; dashed lines show cumulative samples.}
\label{fig:recall}
\par\vspace{12pt}
\subsection*{Why Answer Disagreement for Routing?}
\noindent\begin{minipage}[t]{\columnwidth}\vspace{0pt}
We use persistent disagreement because it directly tracks whether additional sampling recovers agreement under the current problem formulation. Other training-free signals, such as token-level entropy and perplexity, offer alternative measures of uncertainty. However, both depend on model probabilities, which may be unavailable through black-box interfaces. Using these scores for routing also requires a numerical decision threshold that may need calibration across models and tasks.
\end{minipage}\hfill
\begin{minipage}[t]{\columnwidth}\vspace{0pt}
Our sequential checks produce a binary outcome of agreement or disagreement at each step, enabling routing without separately calibrating a continuous uncertainty score. Moreover, the sampled answers serve both strategy selection and final answering, avoiding a separate stability-estimation stage. This provides a lightweight implementation of stability-guided routing without estimating the full answer distribution.
\end{minipage}\par

}]
\clearpage
\twocolumn[{
\noindent\begin{minipage}[t]{\columnwidth}\vspace{0pt}
\subsection{Controlled Analysis of Disagreement-Guided Routing}
\label{app:routing_analysis}
We first report the exact route-level statistics from one representative end-to-end run with seed 42 in Table~\ref{tab:route_statistics}. The overall rows aggregate the final predictions produced by the complete sequential routing procedure. Using the same candidate outputs, we further construct a matched random-routing baseline by preserving the numbers of Check 1, Check 2, and Persistent assignments while randomly permuting these assignments across instances. Each group receives the same strategy and inference budget, so the only difference is the routing rule. As shown in Table~\ref{tab:random_routing}, disagreement-guided routing improves average accuracy by 6.7 points under matched route proportions and generation budgets.

\par\vspace{10pt}
\begingroup
\small
\setlength{\tabcolsep}{2.5pt}
\renewcommand{\arraystretch}{1.08}
\captionsetup{skip=10pt}
\centering
\sbox{\appmeasurebox}{\begin{tabular}{llrrr}
\toprule
Dataset & Route & Instances & Correct & Accuracy \\
\midrule
Math500 & Check 1 & 412 & 399 & 96.8 \\
        & Check 2 & 25  & 20  & 80.0 \\
        & Persistent & 63  & 40  & 63.5 \\
        & Overall & 500 & 459 & 91.8 \\
\midrule
Gaokao EN & Check 1 & 316 & 278 & 88.0 \\
             & Check 2 & 19  & 14  & 73.7 \\
             & Persistent & 50  & 28  & 56.0 \\
             & Overall & 385 & 320 & 83.1 \\
\midrule
Olympiad & Check 1 & 475 & 347 & 73.1 \\
              & Check 2 & 54  & 23  & 42.6 \\
              & Persistent & 146 & 56  & 38.4 \\
              & Overall & 675 & 426 & 63.1 \\
\bottomrule
\end{tabular}}
\setlength{\tabcolsep}{\dimexpr\tabcolsep+(\linewidth-\wd\appmeasurebox)/10\relax}
\begin{tabular}{llrrr}
\toprule
Dataset & Route & Instances & Correct & Accuracy \\
\midrule
Math500 & Check 1 & 412 & 399 & 96.8 \\
        & Check 2 & 25  & 20  & 80.0 \\
        & Persistent & 63  & 40  & 63.5 \\
        & Overall & 500 & 459 & 91.8 \\
\midrule
Gaokao EN & Check 1 & 316 & 278 & 88.0 \\
             & Check 2 & 19  & 14  & 73.7 \\
             & Persistent & 50  & 28  & 56.0 \\
             & Overall & 385 & 320 & 83.1 \\
\midrule
Olympiad & Check 1 & 475 & 347 & 73.1 \\
              & Check 2 & 54  & 23  & 42.6 \\
              & Persistent & 146 & 56  & 38.4 \\
              & Overall & 675 & 426 & 63.1 \\
\bottomrule
\end{tabular}
\par
\captionof{table}{Route-level statistics for the seed-42 end-to-end run.}
\label{tab:route_statistics}
\endgroup
\par\vspace{10pt}
\par\vspace{12pt}
\begingroup
\small
\setlength{\tabcolsep}{2.5pt}
\renewcommand{\arraystretch}{1.08}
\captionsetup{skip=10pt}
\centering
\sbox{\appmeasurebox}{\begin{tabular}{lcccc}
\toprule
Routing rule & Math500 & Gaokao En & Olympiad & Avg. \\
\midrule
Random & 86.3 & 80.0 & 51.5 & 72.6 \\
Ours & \textbf{91.8} & \textbf{83.1} & \textbf{63.1} & \textbf{79.3} \\
\bottomrule
\end{tabular}}
\setlength{\tabcolsep}{\dimexpr\tabcolsep+(\linewidth-\wd\appmeasurebox)/10\relax}
\begin{tabular}{lcccc}
\toprule
Routing rule & Math500 & Gaokao En & Olympiad & Avg. \\
\midrule
Random & 86.3 & 80.0 & 51.5 & 72.6 \\
Ours & \textbf{91.8} & \textbf{83.1} & \textbf{63.1} & \textbf{79.3} \\
\bottomrule
\end{tabular}
\par
\captionof{table}{Accuracy (\%) of random and disagreement-guided routing under matched route proportions and total generation counts.}
\label{tab:random_routing}
\endgroup
\par\vspace{10pt}
\paragraph{Budget accounting.} The matching constraint concerns the total number of base-model generations and the route proportions. Token usage and wall-clock time are separate cost measurements, reported in the main efficiency evaluation.
\end{minipage}\hfill
\begin{minipage}[t]{\columnwidth}\vspace{0pt}
\subsection{Robustness to the Rewriting Prompt}
\label{app:rewrite_prompt}
To examine whether rewriting benefits depend on shortening the problem statement, we evaluate a phrasing-preserving prompt on the same OlympiadBench Persistent instances:

\begin{list}{}{\leftmargin=8pt\rightmargin=8pt}\item[]

\textit{Please rephrase the following question by reorganizing its structure and making any implicit conditions explicit, without shortening its length or omitting any details from the original. Only provide the revised question without answers or calculations.}
\end{list}

As shown in Table~\ref{tab:rewrite_prompt}, the phrasing-preserving prompt retains a similar effective-rewriting rate while reducing the harmful-rewriting rate from 2.8\% to 1.4\%. This indicates that rewriting can remain effective without compressing the original problem statement.

\par\vspace{10pt}
\begingroup
\small
\setlength{\tabcolsep}{2.5pt}
\renewcommand{\arraystretch}{1.08}
\captionsetup{skip=10pt}
\centering
\sbox{\appmeasurebox}{\begin{tabular}{lcc}
\toprule
Rewrite prompt & Effective (\%) & Harmful (\%) \\
\midrule
Original simplifying & \textbf{31.0} & 2.8 \\
Phrasing-preserving  & 29.4 & \textbf{1.4} \\
\bottomrule
\end{tabular}}
\setlength{\tabcolsep}{\dimexpr\tabcolsep+(\linewidth-\wd\appmeasurebox)/6\relax}
\begin{tabular}{lcc}
\toprule
Rewrite prompt & Effective (\%) & Harmful (\%) \\
\midrule
Original simplifying & \textbf{31.0} & 2.8 \\
Phrasing-preserving  & 29.4 & \textbf{1.4} \\
\bottomrule
\end{tabular}
\par
\captionof{table}{Effect of different rewriting prompts on OlympiadBench Persistent instances.}
\label{tab:rewrite_prompt}
\endgroup
\par\vspace{10pt}
\par\vspace{14pt}
\subsection{Comparison with the Original-Formulation Oracle}
\begingroup
\small
\setlength{\tabcolsep}{2.5pt}
\renewcommand{\arraystretch}{1.08}
\captionsetup{skip=10pt}
\centering
\sbox{\appmeasurebox}{\begin{tabular}{lccc}
\toprule
\textbf{Method} & \textbf{Math500} & \textbf{Gaokao En} & \textbf{Olympiad} \\
\midrule
Majority        & 87.6 & 79.7 & 50.7 \\
Pass@6 (oracle) & 91.6 & 82.6 & 59.5 \\
\rowcolor{oursbg}
Ours & \textbf{92.8} & \textbf{82.9} & \textbf{61.9} \\
\bottomrule
\end{tabular}}
\setlength{\tabcolsep}{\dimexpr\tabcolsep+(\linewidth-\wd\appmeasurebox)/8\relax}
\begin{tabular}{lccc}
\toprule
\textbf{Method} & \textbf{Math500} & \textbf{Gaokao En} & \textbf{Olympiad} \\
\midrule
Majority        & 87.6 & 79.7 & 50.7 \\
Pass@6 (oracle) & 91.6 & 82.6 & 59.5 \\
\rowcolor{oursbg}
Ours & \textbf{92.8} & \textbf{82.9} & \textbf{61.9} \\
\bottomrule
\end{tabular}
\par
\captionof{table}{Accuracy (\%) using Qwen3-8B with an 8K-token generation limit per call. The empirical pass@6 oracle uses the same six original-formulation candidates as Majority, counting an instance as correct if any candidate is correct.}
\label{tab:pass6_oracle}
\endgroup
\par\vspace{10pt}
The accuracy margins over this oracle are 1.2, 0.3, and 2.4 percentage points on Math500, Gaokao En, and Olympiad, respectively. These results compare selective reformulation with oracle selection from original-formulation candidates.
\end{minipage}\par

}]
\clearpage
\twocolumn[{
\section{Prompt Templates}
\label{appendix:prompts}
\begin{tcolorbox}[
    enhanced,
    title=Prompt Templates,
    fonttitle=\bfseries,
    fontupper=\normalsize,
    colback=white,
    before upper={%
        \setlength{\abovedisplayskip}{1pt}%
        \setlength{\belowdisplayskip}{1pt}%
        \setlength{\abovedisplayshortskip}{1pt}%
        \setlength{\belowdisplayshortskip}{1pt}%
    },
    width=\textwidth, % 占满右半栏
]

\textbf{Mathematical Reasoning Prompt:}\\
Please reason step by step, and put your final answer within boxed\{\}.

\medskip
\hrule height 0.6pt
\vspace{6pt}

\textbf{Rewriting Prompt for Mathematical Problems:}\\
Please remove unnecessary descriptions from the following question, simplify its length while keeping the original meaning unchanged, and retain important numbers and symbols. Only provide the revised question without answers or calculations.

\medskip
\hrule height 0.6pt
\vspace{6pt}

\textbf{Code Generation Prompt:}\\
Please provide a self-contained Python script that solves the following problem in a markdown code block: \{\textit{input}\} Below is a Python script with a self-contained function that solves the problem and passes corresponding tests:

\medskip
\hrule height 0.6pt
\vspace{6pt}

\textbf{Rewriting Prompt for Coding Problems:}\\
Rewrite the following programming problem in your own words. Keep the meaning, constraints, and examples exactly the same, but rephrase the description naturally. Do NOT provide any solution or any code besides the rewritten problem text. \{\textit{input}\} Here is the rewritten problem:

\end{tcolorbox}\par\vspace{16pt}
\section{Answer Equivalence Checking}
\label{sec:equivalence}
\noindent\begin{minipage}[t]{\columnwidth}\vspace{0pt}
To determine whether two answers are equivalent, we adopt a hierarchical matching procedure that goes beyond exact string matching.

\paragraph{Normalization.}
We first normalize both prediction and reference by removing formatting artifacts such as extra whitespace, brackets, and casing differences. Structured outputs are converted into consistent canonical forms.

\paragraph{Numerical Equivalence.}
When both answers can be parsed as numeric values, we perform tolerance-based comparison. We additionally account for common variations such as percentage and scale transformations (e.g., $x$, $x/100$, and $100x$).

\paragraph{Routing usage.}
During routing, matching is applied between generated candidates. Final-answer correctness is evaluated separately against the benchmark references.
\end{minipage}\hfill
\begin{minipage}[t]{\columnwidth}\vspace{0pt}
\paragraph{Structural Matching.} 
For structured outputs such as tuples, lists, and matrices, we perform element-wise comparison recursively.

\paragraph{Symbolic Equivalence.} 
When numerical comparison is not applicable, we apply symbolic matching. In particular, equations are first normalized into residual forms (e.g., $x = y$ to $x - y = 0$) before symbolic equivalence checking.

This hierarchical procedure allows us to capture a broad class of mathematically equivalent answers beyond strict string matching, while remaining efficient in practice.

\paragraph{Code-generation agreement.}
For code generation, agreement is assessed by executing candidate programs on the example inputs provided in the problem statement. Matching outputs on all available examples provides the agreement signal for the same sequential policy. Benchmark accuracy is evaluated separately using EvalPlus.
\end{minipage}\par

}]
\clearpage

\section{Case Study}
\label{sec:appendix:case_study}

\noindent
\fontsize{9pt}{10pt}\selectfont
\begin{minipage}{\textwidth}
\renewcommand{\arraystretch}{1.25}
\begin{tabularx}{\textwidth}{>{\centering\arraybackslash}p{2cm} >{\centering\arraybackslash}X >{\centering\arraybackslash}X}
\toprule
\textbf{Case} & \textbf{Original} & \textbf{Rewritten} \\
\midrule

hidden info error
&
\begin{minipage}[t]{\linewidth}
\textbf{Question:} Determine the smallest positive real number $k$ with the following property. Let $A B C D$ be a convex quadrilateral, and let points $A_{1}, B_{1}, C_{1}$ and $D_{1}$ lie on sides $A B, B C, C D$ and $D A$, respectively. Consider the areas of triangles $A A_{1} D_{1}, B B_{1} A_{1}, C C_{1} B_{1}$, and $D D_{1} C_{1}$; let $S$ be the sum of the two smallest ones, and let $S_{1}$ be the area of quadrilateral $A_{1} B_{1} C_{1} D_{1}$. Then we always have $k S_{1}\geq S$.
\textbf{Answer:} $1$

\textbf{Output:} $\cdots$ \textcolor{red}{By choosing points $A_{1}, B_{1}, C_{1}$ and $D_{1}$ arbitrarily close to the vertices of the quadrilateral}, the area $S_{1}$ of $A_{1} B_{1} C_{1} D_{1}$ can be made arbitrarily small $\cdots$ Final Answer $\boxed{\frac{1}{2}}$
\end{minipage}
&
\begin{minipage}[t]{\linewidth}
\textbf{Question:} Let $ABCD$ be a convex quadrilateral. Choose points $A_1 \in AB, B_1 \in BC, C_1 \in CD, D_1 \in DA$ strictly inside each side so that $A_1B_1C_1D_1$ is convex and lies entirely inside $ABCD$.
Define
$
T_1=[AA_1D_1],\quad T_2=[BB_1A_1],\quad T_3=[CC_1B_1],\quad T_4=[DD_1C_1]
$
and let $S$ be the sum of the two smallest of these areas. Let $S_1=[A_1B_1C_1D_1]$. 
Find the smallest $k>0$ such that
$
k\,S_1 \ge S
$
holds for all configurations.
\textbf{Answer:} $1$

\textbf{Output:} Let $ABCD$ be convex, and points $A_1, B_1, C_1, D_1$ chosen cyclically. 
\textcolor{OliveGreen}{Then $A_1B_1C_1D_1$ is always convex inside $ABCD$} $\cdots$ Final Answer $\boxed{1}$
\end{minipage}
\\
\midrule

answer format error
&
\begin{minipage}[t]{\linewidth}
\textbf{Question:} Given sets $M=\{x|x+2\geq 0\},N=\{x|x-1<0\}$, find $M\cap N$.
\textbf{Answer:} $\{x|-2\leq x < 1\}$

\textbf{Output:} To find the $\cdots$ this is: $[-2, 1)$ \textcolor{red}{Final Answer $\boxed{[-2, 1)}$}
\end{minipage}
&
\begin{minipage}[t]{\linewidth}
\textbf{Question:} $M=\{x|x+2\geq 0\},N=\{x|x-1<0\}$, find $M\cap N$.
\textbf{Answer:} $\{x|-2\leq x < 1\}$

\textbf{Output:} To find the $\cdots$ but not including, \textcolor{red}{Final Answer$\boxed{[-2, 1)}$}
\end{minipage}
\\
\midrule

context loss error
&
\begin{minipage}[t]{\linewidth}
\textbf{Question:} Let $A$ be the sum of the digits of the number you will receive from position $7$, and let $B$ be the sum of the digits of the number you will receive from position $9$. Let $(x, y)$ be a point randomly selected from the interior of the triangle whose consecutive vertices are $(1,1),(B, 7)$ and $(17,1)$. Compute the probability that $x>A-1$. 
\textbf{Answer:} $\frac{79}{128}$

\textbf{Output:} Let the triangle $\cdots$ For a fixed $y$, let $x_{min}=1$ $\cdots$ Since $x$ ranges from $1$ to $17$, \textcolor{red}{we compute $P(x>A-1)=\frac{17-(A-1)}{16}$} $\cdots$ 
\end{minipage}
&
\begin{minipage}[t]{\linewidth}
\textbf{Question:} Compute the probability that $x>A-1$ where 
$A$ is the sum of the digits of the number from position $7$ and $B$ is the sum of the digits of the number from position $9$, with $(x, y)$ selected randomly from the interior of the triangle with vertices $(1,1),(B, 7)$ and $(17,1)$.
\textbf{Answer:} $\frac{79}{128}$

\textbf{Output:} Let $(x,y)$ be uniformly $\cdots$ The area of the triangle is $S=\frac{1}{2}(17-1)(7-1)=48$ $\cdots$ \textcolor{red}{we compute $P(x>A-1)=\frac{17-(A-1)}{16}$} $\cdots$
\end{minipage}
\\
\midrule

calculation error
&
\begin{minipage}[t]{\linewidth}
\textbf{Question:} Determine the minimum potential in V (to 2 decimal places) that must be applied to an $\alpha$-particle so that on interaction with a hydrogen atom, a ground state electron will be excited to $n$ $=6$.
\textbf{Answer:} $6.62$

\textbf{Output:} $\cdots$ Sixth energy level ($ n = 6 $): $  E_6 = -\frac{13.6}{36} \approx -0.3778 \text{eV}$. \textcolor{red}{Energy difference required to excite the electron $\Delta E = E_6 - E_1 = -0.3778 - (-13.6) = 13.2222 \text{eV}$} $\cdots$ Solving for V: $V = 13.22 / 2 = 6.61 V$. Final Answer $\boxed{6.61}$
\end{minipage} 
&
\begin{minipage}[t]{\linewidth}
\textbf{Question:} Determine the minimum electric potential (rounded to 2 decimal places) that must be applied to an $\alpha$-particle (charge $+2e$) so that, when it interacts with a hydrogen atom, it transfers enough energy to excite the hydrogen atom’s electron from the ground state ($n=1$) to the sixth energy level ($n=6$).
\textbf{Answer:} $6.62$

\textbf{Output:} $\cdots$ Determine the Energy Required for Excitation. \textcolor{red}{The energy levels of a hydrogen atom are given by the formula: $E_n = -\frac{13.6}{n^2}\text{ eV}$ For the ground state $ n = 1 $: $E_1 = -13.6\text{ eV}$} $\cdots$ Final Answer $\boxed{6.61}$ 
\end{minipage}
\\
\midrule

over-compression error
&
\begin{minipage}[t]{\linewidth}
\textbf{Question:} 50 people sign up for a football club, 60 people sign up for a table tennis club, and 70 people sign up for either the football or the table tennis club. There are no other clubs and everyone signs up for at least one of the clubs. If someone signs up for the football club, what is the probability that the same person signs up for the table tennis club?
\textbf{Answer:} $0.8$

\textbf{Output:} \textcolor{OliveGreen}{To solve this problem, we are given the following information: 50 people signed up for the football club. 60 people signed up for the table tennis club. 70 people signed up for either the football or the table tennis club.} $\cdots$ Final Answer $\boxed{0.8}$
\end{minipage}
&
\begin{minipage}[t]{\linewidth}
\textbf{Question:} If someone signs up for the football club, what is the probability they also signed up for the table tennis club?
\textbf{Answer:} $0.8$

\textbf{Output:} To determine the probability that someone who signed up for the football club also signed up for the table tennis club, we are dealing with a conditional probability problem. $\cdots$ \textcolor{red}{However, no specific data or probabilities are provided in the question. Without knowing} $\cdots$ \textcolor{red}{Hypothetical Scenario If we assume a common textbook-type scenario, for example: Total of 100 people, 60 signed up for football, 20 signed up for both football and table tennis} $\cdots$ the probability would be: $\boxed{\dfrac{1}{3}}$
\end{minipage}
\\

\bottomrule
\end{tabularx}
\end{minipage}

\begin{figure*}[t]
\centering
\begin{tcolorbox}[
    enhanced,
    title=Case 1 (Original),
    fonttitle=\bfseries,
    colback=white,
    before upper={%
        \setlength{\abovedisplayskip}{1pt}%
        \setlength{\belowdisplayskip}{1pt}%
        \setlength{\abovedisplayshortskip}{1pt}%
        \setlength{\belowdisplayshortskip}{1pt}%
    },
]
% ---------- Question ----------
\textbf{Original Question:}\\
Suppose the polynomial
$$
f(x) = a_nx^n + a_{n-1}x^{n-1} + \cdots + a_1x + a_0
$$
has integer coefficients and distinct integer roots.  
Given $a_n = 2$ and $a_0 = 66$, find the least possible value of $|a_{n-1}|$.\\[4pt]
\hrule height 0.6pt
\vspace{6pt}
% ---------- Model Output ----------
\textbf{Model Output:}\\
To solve the problem, we are given a polynomial of the form:$$f(x) = a_nx^n + a_{n-1}x^{n-1} + \cdots + a_0$$ with integer coefficients, distinct integer roots, and the following constraints:$ a_n = 2 $ $ a_0 = 66 $ We are to find the least possible value of $ |a_{n-1}| $ Step 1: Use Vieta's Formulas.
The polynomial can be written in factored form as:
$$f(x) = 2(x - r_1)(x - r_2)\cdots(x - r_n)$$
where $ r_1, r_2, \ldots, r_n $ are distinct integers (the roots), and $ a_n = 2 $.
The constant term $ a_0 $ is given by:
$$a_0 = 2 \cdot (-1)^n \cdot r_1 r_2 \cdots r_n$$
We are told $ a_0 = 66 $, so:
$2 \cdot (-1)^n \cdot (r_1 \cdots r_n) = 66 \Rightarrow (-1)^n \cdot (r_1 \cdots r_n) = 33$
Thus, the product of the roots (with sign) is:
$$r_1 r_2 \cdots r_n = \pm 33$$
Step 2: Express $ a_{n-1} $ in Terms of Roots.
From Vieta's formulas, the coefficient $ a_{n-1} $ is given by: $$a_{n-1} = -2 \cdot \text{(sum of roots)}$$
So:$$|a_{n-1}| = 2 \cdot |\text{sum of roots}|$$
Our goal is to minimize this value, which reduces to finding a set of distinct integers whose product is $ \pm 33 $, and whose sum has the smallest absolute value.
Step 3: Try Different Degrees $ n $.
Let’s consider different possible values of $ n $, and for each, find the smallest possible absolute value of the sum of distinct integer roots with product $ \pm 33 $. 
Case $ n = 4 $: We want four distinct integers whose product is $ 33 $ (since $ n = 4 $ is even, the product is positive). Let’s try: $ r_1 = 1, r_2 = -1, r_3 = 3, r_4 = -11 $ Check product:$$1 \cdot (-1) \cdot 3 \cdot (-11) = 33 \quad$$ Sum:$$1 + (-1) + 3 + (-11) = -8 \Rightarrow |a_{n-1}| = 2 \cdot 8 = 16$$
This is a candidate for the minimum. Try other combinations: 
$r_1 = -1$, $r_2 = 1$, $r_3 = -3$, $r_4 = 11$; product $= 33$, sum $= -1 + 1 - 3 + 11 = 8$, same result. 
Thus, minimum $|a_{n-1}| = 16$ for $n = 4$.
Check for smaller $|a_{n-1}|$:
Try other values of $ n $:
\begin{itemize}
    \item $n = 2$: roots $(3, 11)$, sum $= 14 \Rightarrow |a_{n-1}| = 28$
    \textcolor{red}{\item $n = 3$: roots $(-3, 1, 11)$, sum $= 9 \Rightarrow |a_{n-1}| = 18$}
    \item $n = 1$: root $-33 \Rightarrow |a_{n-1}| = 66$
    \item $n = 5$: no valid set of 5 distinct integers with product $-33$
    \item $n = 6$: no valid set of 6 distinct integers with product $33$
\end{itemize}
\textcolor{red}{So, the minimal value occurs at $n = 4$} with $|a_{n-1}| = 16$. Final Answer $
\boxed{16}
$
\end{tcolorbox}
\end{figure*}

\begin{figure*}[t]
\centering
\begin{tcolorbox}[
    enhanced,
    title=Case 1 (Rewritten),
    fonttitle=\bfseries,
    colback=white,
    before upper={%
        \setlength{\abovedisplayskip}{1pt}%
        \setlength{\belowdisplayskip}{1pt}%
        \setlength{\abovedisplayshortskip}{1pt}%
        \setlength{\belowdisplayshortskip}{1pt}%
    },
]

% ---------- Question ----------
\textbf{Rewritten Question:}\\
Suppose the polynomial $$f(x) = 2x^n + a_{n-1} x^{n-1} + \cdots + a_0$$ has distinct integer roots and $a_0 = 66$. What is the least possible value of $|a_{n-1}|$?\\[4pt]
\hrule height 0.6pt
\vspace{6pt}

% ---------- Model Output ----------
\textbf{Model Output:}\\
To determine the least possible value of $ |a_{n-1}| $ for a polynomial with integer coefficients, distinct integer roots, leading coefficient $ a_n = 2 $, and constant term $ a_0 = 66 $, we proceed as follows: Step 1: Polynomial Structure.
The polynomial is of the form: $$f(x) = 2(x - r_1)(x - r_2)\cdots(x - r_n)$$ where $ r_1, r_2,\dots, r_n $ are distinct integers. The leading coefficient is 2, and the constant term $ a_0 $ is given as 66. Step 2: Constant Term and Product of Roots. Using the expansion of the polynomial, the constant term is:$$a_0 = 2\cdot (-1)^n\cdot r_1 r_2\cdots r_n$$ Given $ a_0 = 66 $, we solve:$$2\cdot (-1)^n\cdot (r_1 r_2\cdots r_n) = 66\Rightarrow r_1 r_2\cdots r_n =\frac{66}{2\cdot (-1)^n} =\frac{33}{(-1)^n}$$ Thus, the product of the roots is $\pm 33 $, and the roots must be distinct integers. Step 3: Coefficient $ a_{n-1} $ and Its Absolute Value. Using Vieta’s formula, the coefficient $ a_{n-1} $ is related to the sum of the roots: $$a_{n-1} = -2\cdot (r_1 + r_2 +\cdots + r_n)$$ So, the absolute value is:$$|a_{n-1}| = 2\cdot |r_1 + r_2 +\cdots + r_n|$$To minimize $ |a_{n-1}| $, we must minimize the absolute value of the sum of the roots, subject to the constraint that the product of the roots is $\pm 33 $, and the roots are distinct integers. Step 4: Searching for Optimal Root Combinations. We need to find distinct integers whose product is $\pm 33 $ and whose sum has the smallest absolute value. \textcolor{OliveGreen}{The integer divisors of 33 are:$$\pm 1,\pm 3,\pm 11,\pm 33$$ 
We test combinations of three distinct integers(since more than three would require duplicate factors or make the sum too large) that multiply to $\pm 33 $. Among the valid combinations, the one that minimizes $ |r_1 + r_2 + r_3| $ is: $ r_1 = 1 $, $ r_2 = 3 $, $ r_3 = -11 $ Product: $ 1\cdot 3\cdot (-11) = -33 $ Sum: $ 1 + 3 - 11 = -7 $\ Absolute value of sum: $ 7 $
This yields:$$|a_{n-1}| = 2\cdot | -7 | = 14$$} Other combinations (e.g., $ -1, -3, 11 $) also yield the same minimal sum of 7 in absolute value. Step 5: Verifying Optimality. \textcolor{OliveGreen}{Trying all other combinations of 2 or 4 distinct integers with product $\pm 33 $ results in larger absolute sums. For example: \[
\begin{aligned}
&(3, -11): && \text{sum} = -8, && |a_{n-1}| = 16,\\
&(1, -1, 3, -11): && \text{sum} = -8, && |a_{n-1}| = 16,\\
&(1, -1, -3, 11): && \text{sum} = 8, && |a_{n-1}| = 16.
\end{aligned}
\] All of these are greater than 14.} Hence, the minimum possible value of $ |a_{n-1}| $ is achieved with the root set $ (1, 3, -11) $ or $ (-1, -3, 11) $, both giving a minimal sum of 7 in absolute value. Final Answer $\boxed{14}$
\end{tcolorbox}
\end{figure*}

\begin{figure*}[t]
\centering
\begin{tcolorbox}[
    enhanced,
    title=Case 2 (Original),
    fonttitle=\bfseries,
    colback=white,
    before upper={%
        \setlength{\abovedisplayskip}{1pt}%
        \setlength{\belowdisplayskip}{1pt}%
        \setlength{\abovedisplayshortskip}{1pt}%
        \setlength{\belowdisplayshortskip}{1pt}%
    },
]
% ---------- Question ----------
\textbf{Original Question:}\\
An integer $n \geqslant 3$ is given. We call an $n$-tuple of real numbers
$\left(x_{1}, x_{2}, \ldots, x_{n}\right)$ \emph{Shiny} if for each permutation
$y_{1}, y_{2}, \ldots, y_{n}$ of these numbers we have
$
\sum_{i=1}^{n-1} y_i y_{i+1} 
= y_1 y_2 + y_2 y_3 + \cdots + y_{n-1} y_n 
\geqslant -1.
$
Find the largest constant $K = K(n)$ such that
$
\sum_{1 \leqslant i < j \leqslant n} x_i x_j \geqslant K
$
holds for every Shiny $n$-tuple $\left(x_{1}, x_{2}, \ldots, x_{n}\right)$.
\\[4pt]
\hrule height 0.6pt
\vspace{6pt}
% ---------- Model Output ----------
\textbf{Model Output:}\\
To solve the problem, we are given an integer $ n \geq 3 $, and an $ n $-tuple $ (x_1, x_2, \cdots, x_n) $ is called Shiny if for every permutation $ (y_1, y_2, \cdots, y_n) $ of the tuple, the sum $$\sum_{i=1}^{n-1} y_i y_{i+1} \geq -1$$ holds. Our goal is to find the largest constant $ K(n) $ such that for every Shiny $ n $-tuple, the sum of all pairwise products$$\sum_{1 \leq i < j \leq n} x_i x_j$$ is at least $ K(n) $. Step 1: Understanding the Objective. We are asked to find the minimum value of the expression:$$\sum_{1 \leq i < j \leq n} x_i x_j$$ over all Shiny $ n $-tuples. Then, the largest constant $ K(n) $ such that this sum is always at least $ K(n) $ is precisely that minimum value. Step 2: Key Observations. We can express the sum of all pairwise products as: $$\sum_{1 \leq i < j \leq n} x_i x_j = \frac{1}{2} \left( \left( \sum_{i=1}^n x_i \right)^2 - \sum_{i=1}^n x_i^2 \right)$$ However, this expression is not immediately helpful for our purpose. Instead, we focus on constructing examples of Shiny $ n $-tuples and computing the value of the pairwise sum. Step 3: Constructing a Shiny $ n $-tuple with Minimal Pairwise Sum.
Let us consider the $ n $-tuple:$$(x_1, x_2, \ldots, x_n) = (a, -\frac{1}{a}, 0, 0, \ldots, 0)$$ Then, the pairwise sum becomes: $$\sum_{1 \leq i < j \leq n} x_i x_j = a \cdot \left( -\frac{1}{a} \right) = -1$$ Now, we must verify that this tuple is Shiny, i.e., that for every permutation of the tuple, the sum of adjacent products is at least $ -1 $. In any permutation, the only non-zero product will be between $ a $ and $ -\frac{1}{a} $, which is $ -1 $. All other products will be between $ a $ and $ 0 $, or between $ -\frac{1}{a} $ and $ 0 $, or between $ 0 $ and $ 0 $, all of which are $ 0 $. Therefore, the sum of adjacent products in any permutation is either $ -1 $ (when the two non-zero elements are adjacent) or $ 0 $ (otherwise). Thus, the minimum sum over all permutations is $ -1 $, which satisfies the Shiny condition. Step 4: Can We Do Better? \textcolor{red}{Suppose there exists a Shiny $ n $-tuple with pairwise sum less than $ -1 $. Let’s assume that such a tuple exists. Then, the minimum sum over all permutations of adjacent products would be less than $ -1 $, which would violate the Shiny condition. Therefore, such a tuple cannot exist. This implies that the minimal value of the pairwise sum over all Shiny $ n $-tuples is exactly $ -1 $.} Step 5: Conclusion. We have constructed a Shiny $ n $-tuple with pairwise sum $ -1 $, and we have shown that it is not possible to construct a Shiny $ n $-tuple with a pairwise sum less than $ -1 $. Therefore, the largest constant $ K(n) $ such that the pairwise sum is always at least $ K(n) $ for all Shiny $ n $-tuples is: $\boxed{-1}$
\end{tcolorbox}
\end{figure*}

\begin{figure*}[t]
\centering
\begin{tcolorbox}[
    enhanced,
    title=Case 2 (Rewritten),
    fonttitle=\bfseries,
    colback=white,
    before upper={%
        \setlength{\abovedisplayskip}{1pt}%
        \setlength{\belowdisplayskip}{1pt}%
        \setlength{\abovedisplayshortskip}{1pt}%
        \setlength{\belowdisplayshortskip}{1pt}%
    },
]

% ---------- Question ----------
\textbf{Rewritten Question:}\\
An integer $n\ge3$ is given. We call an $n$-tuple of real numbers 
$(x_1, x_2, \ldots, x_n)$ \emph{Shiny} if for every permutation 
$(y_1, y_2, \ldots, y_n)$ of these numbers, the following inequality holds:
$
y_1 y_2 + y_2 y_3 + \cdots + y_{n-1} y_n \ge -1.
$
For every Shiny $n$-tuple, consider the sum of all pairwise products
$
S = \sum_{1 \le i < j \le n} x_i x_j.
$
Find the largest constant $K(n)$ such that
$
S \ge K(n)
$
holds for all Shiny $n$-tuples.
\\[4pt]
\hrule height 0.6pt
\vspace{6pt}

% ---------- Model Output ----------
\textbf{Model Output:}\\
To determine the largest constant $ K(n) $ such that for every Shiny $ n $-tuple, the sum of all products of pairs $ x_i x_j $ for $ i < j $ is at least $ K(n) $, we begin by understanding the key constraints imposed by the Shiny condition. Understanding the Shiny Condition. A tuple $ (x_1, x_2, \dots, x_n) $ is Shiny if, for any permutation of the tuple, the sum of adjacent products is at least $-1$. That is: $$\min_{\text{permutations} \pi} \left( \sum_{i=1}^{n-1} \pi_i \pi_{i+1} \right) \ge -1$$ This condition ensures that no matter how we rearrange the elements, the sum of adjacent products never drops below $-1$. Goal: Find the Largest Lower Bound $ K(n) $. We are to find the largest constant $ K(n) $ such that for every Shiny $ n $-tuple, the sum of all pairwise products is at least $ K(n) $, that is: $$\sum_{1 \le i < j \le n} x_i x_j \ge K(n)$$ \textcolor{OliveGreen}{We can relate this sum to the square of the sum of the elements: $$\left( \sum_{i=1}^n x_i \right)^2 = \sum_{i=1}^n x_i^2 + 2 \sum_{1 \le i < j \le n} x_i x_j$$} Rearranging gives: $$\sum_{1 \le i < j \le n} x_i x_j = \frac{1}{2} \left( \left( \sum x_i \right)^2 - \sum x_i^2 \right)$$ This expression will be key in analyzing the behavior of the sum of pairwise products.
Constructing a Candidate Shiny Tuple. \textcolor{OliveGreen}{To find the minimum value of $ \sum x_i x_j $ across all Shiny $ n $-tuples, we consider a specific configuration: Let $ n-1 $ elements be $ \varepsilon $ (a small positive number) Let the last element be $ -c $, where $ c > 0 $.} This gives us a tuple of the form $ (\varepsilon, \varepsilon, \dots, \varepsilon, -c) $, with $ n-1 $ copies of $ \varepsilon $ and one $ -c $. We analyze the sum of pairwise products: $$\sum_{1 \le i < j \le n} x_i x_j = \frac{1}{2} \left( \left( (n-1)\varepsilon - c \right)^2 - \left( (n-1)\varepsilon^2 + c^2 \right) \right)$$ After simplification, this becomes: $\sum x_i x_j = \frac{n-1}{2} \left( \varepsilon^2 - 1 \right)$ To ensure the tuple is Shiny, we must also satisfy the condition that the minimum sum of adjacent products over all permutations is at least $-1$. By analyzing the worst-case permutation (when $ -c $ is adjacent to two $ \varepsilon $'s), we find that this condition is satisfied as $ \varepsilon \to 0 $, and we can set $ c $ such that the sum of adjacent products is exactly $-1$. Asymptotic Behavior and Conclusion. As $ \varepsilon \to 0 $, the sum of pairwise products approaches: $\sum x_i x_j \to \frac{n-1}{2} \cdot (-1) = -\frac{n-1}{2}$ This is the smallest possible value of the sum of pairwise products that still satisfies the Shiny condition. Since we can construct such a tuple that approaches this value, and no tuple can have a smaller sum without violating the Shiny condition, we conclude that: $\boxed{-\frac{n-1}{2}}$ is the largest constant $ K(n) $ such that for every Shiny $ n $-tuple, the sum of all products of pairs is at least $ K(n) $. Final Answer $\boxed{-\dfrac{n-1}{2}}$
\end{tcolorbox}
\end{figure*}

\endgroup

\end{document}